\documentclass[10pt,twocolumn,letterpaper]{article}

\usepackage{cvpr}              %

\usepackage{graphicx}
\usepackage{amsmath}
\usepackage{amssymb}
\usepackage{booktabs}
\usepackage{color}
\usepackage{soul}
\usepackage[symbol]{footmisc}
\usepackage[accsupp]{axessibility} %

\usepackage[labelfont=bf]{caption}

\usepackage[breaklinks=true,bookmarks=false,colorlinks=true,linkcolor=red,citecolor=blue]{hyperref}
\usepackage[table,x11names,dvipsnames,table]{xcolor}

\usepackage[capitalize]{cleveref}
\crefname{section}{Sec.}{Secs.}
\Crefname{section}{Section}{Sections}
\Crefname{table}{Table}{Tables}
\crefname{table}{Tab.}{Tabs.}

\definecolor{DarkGreen}{rgb}{0.0, 0.7, 0.5}
\definecolor{traj_red}{rgb}{0.78431373, 0.00784314, 0.09411765}
\definecolor{traj_blue}{rgb}{0.0, 0.02745098, 0.67058824}
\definecolor{traj_green}{rgb}{0.058823529411764705, 0.8117647058823529, 0.2196078431372549}
\definecolor{traj_bright_purple}{rgb}{0.79607843, 0.65098039, 0.79215686}
\definecolor{DarkRed}{rgb}{0.8, 0.0, 0.4}
\definecolor{DarkCyan}{rgb}{0.4, 0.85, 0.8}

\def\cvprPaperID{9301} %
\def\confName{CVPR}
\def\confYear{2023}

\begin{document}

\title{Avatars Grow Legs: Generating Smooth Human Motion \\ from Sparse Tracking Inputs with Diffusion Model}

\author{
\parbox{\linewidth}{\centering
Yuming Du\footnotemark \hspace{12pt}
Robin Kips \hspace{10pt}
Albert Pumarola \hspace{10pt}
Sebastian Starke \hspace{10pt} 
Ali Thabet \hspace{10pt} 
Artsiom Sanakoyeu
}\\
Meta AI
}

\twocolumn[{%
\renewcommand\twocolumn[1][]{#1}
\maketitle
\vspace{-1.1cm}
\begin{center}
    \centering 

    \includegraphics[width=1.0\textwidth]{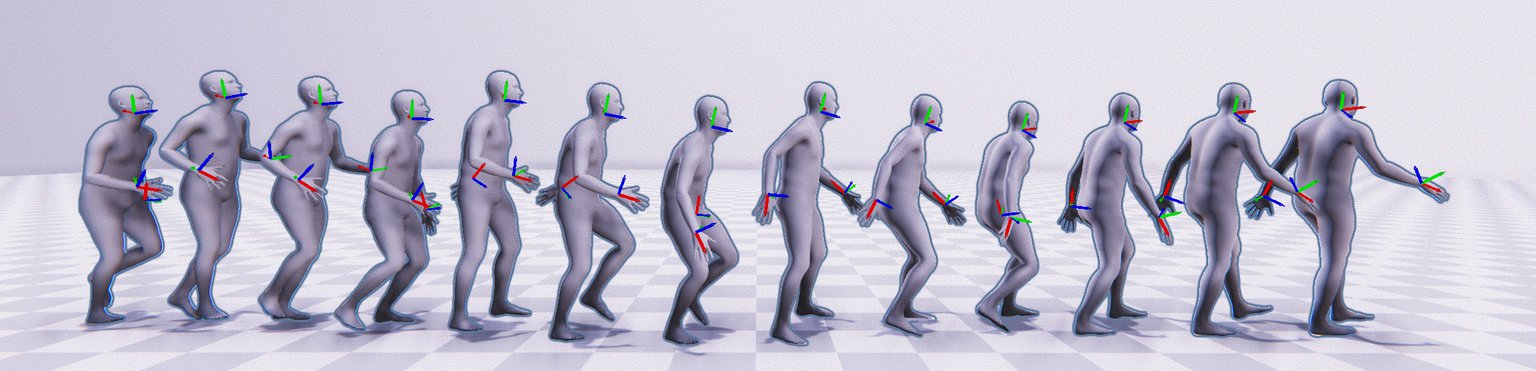}
    \captionof{figure}{\textbf{Full body motion synthesis based on HMD and hand controllers input.} 
    We show synthesis results of the proposed AGRoL method. RGB axes illustrate the %
    orientation of the head and hands which serves as the input to to our model.
    }
    \label{fig:pull_figure}
\end{center}
}]

\footnotetext[1]{Work done during an internship at Meta AI.}
\footnotetext[0]{Code is available at \href{https://github.com/facebookresearch/AGRoL}{github.com/facebookresearch/AGRoL}.}

\begin{abstract}
With the recent surge in popularity of AR/VR applications, realistic and accurate control of 3D full-body avatars has become a highly demanded feature. A particular challenge is that only a sparse tracking signal is available from standalone HMDs (Head Mounted Devices), often limited to tracking the user's head and wrists. While this signal is resourceful for reconstructing the upper body motion, the lower body is not tracked and must be synthesized from the limited information provided by the upper body joints. In this paper, we present AGRoL, a novel conditional diffusion model specifically designed to track full bodies given sparse upper-body tracking signals. Our model is based on a simple multi-layer perceptron (MLP) architecture and a novel conditioning scheme for motion data. It can predict accurate and smooth full-body motion, particularly the challenging lower body movement. Unlike common diffusion architectures, our compact architecture can run in real-time, making it suitable for online body-tracking applications. We train and evaluate our model on AMASS motion capture dataset, and demonstrate that our approach outperforms state-of-the-art methods in generated motion accuracy and smoothness. We further justify our design choices through extensive experiments and ablation studies.

\end{abstract}
\section{Introduction}
\label{sec:intro}

Humans are the primary actors in AR/VR applications. As such, being able to track full-body movement is in high demand for these applications.
Common approaches are able to accurately track upper bodies only~\cite{yang2021lobstr, jiang2022transformer}. 
Moving to full-body tracking
unlocks engaging experiences where users can interact with the virtual environment with an increased sense of presence. 

However, in the typical AR/VR setting there is no strong tracking signal for the entire human body -- only the head and hands are usually tracked by means of Inertial Measurement Unit (IMU) sensors embedded in Head Mounted Displays (HMD) and hand controllers.
Some works suggest adding additional IMUs to track the lower body joints~\cite{huang2018deep, jiang2022transformer}, those additions come at higher costs and the expense of the user's comfort~\cite{kaufmann2021pose,jiang2022avatarposer}. 
In an ideal setting, we want to enable high-fidelity full-body tracking using the standard three inputs (head and hands) provided by most HMDs.

Given the position and orientation information of the head and both hands, predicting full-body pose, especially the lower body, is inherently an underconstrained problem. To address this challenge, different methods rely on generative models such as normalizing flows~\cite{rezende2015variational} and Variational Autoencoders (VAE)~\cite{dittadi2021full_VAE_HMD} to synthesize lower body motions. In the realm of generative models, diffusion models have recently shown impressive results in image and video generation~\cite{sohl2015deep, ho2020denoising, nichol2021improved}, especially for conditional generation. This inspires us to employ the diffusion model to generate the fully-body poses conditioned on the sparse tracking signals. %
To the best of our knowledge, there is no existing work leveraging the diffusion model solely for motion reconstruction from sparse tracking information.

However, it is not trivial to employ the diffusion model in this task.
Existing approaches for conditional generation with diffusion models are widely used for cross-modal conditional generation. Unfortunately, these methods can not be directly applied to the task of motion synthesis, given the disparity in data representations, \eg human body joints feature \vs images.

In this paper, we propose a novel diffusion architecture -- \emph{Avatars Grow Legs} (AGRoL), which is specifically tailored for the task of conditional motion synthesis. Inspired by recent work in future motion prediction~\cite{guo2022simple_mlp}, which uses an MLP-based architecture, we find that a carefully designed MLP network can achieve comparable performance to the state-of-the-art methods. However, we discovered that the predicted motions of MLP networks may contain jittering artifacts. To address this issue and generate smooth realistic full body motion from sparse tracking signals, we design a novel lightweight diffusion model powered by our MLP architecture. Diffusion models require time step embedding~\cite{ho2020denoising, nichol2021glide} to be injected in the network during training and inference; however, we found that our MLP architecture is not sensitive to the positional embedding in the input. To tackle this problem, we propose a novel strategy to effectively inject the time step embedding during the diffusion process. With the proposed strategy, we can significantly mitigate the jittering issues and further improve the model's performance and robustness against the loss of tracking signal. Our model accurately predicts full-body motions, outperforming state-of-the-art methods as demonstrated by the experiments on AMASS~\cite{AMASS:ICCV:2019}, large motion capture dataset.

We summarize our contributions as follows:
\begin{itemize}
    \item We propose AGRoL, a conditional diffusion model specifically designed for full-body motion synthesis based on sparse IMU tracking signals. AGRoL is a simple and yet efficient MLP-based diffusion model with a lightweight architecture. To enable gradual denoising and produce smooth motion sequences we propose a block-wise injection scheme that adds diffusion timestep embedding before every intermediate block of the neural network. With this timestep embedding strategy, AGRoL achieves state-of-the-art performance on the full-body motion synthesis task without any extra losses that are commonly used in other motion prediction methods.
    
    \item We show that our lightweight diffusion-based model AGRoL can generate realistic smooth motions while achieving real-time inference speed, making it suitable for online applications. Moreover, it is more robust against tracking signals loss then existing approaches.
  
\end{itemize}

\begin{figure}
  \centering
    \includegraphics[width=\linewidth]{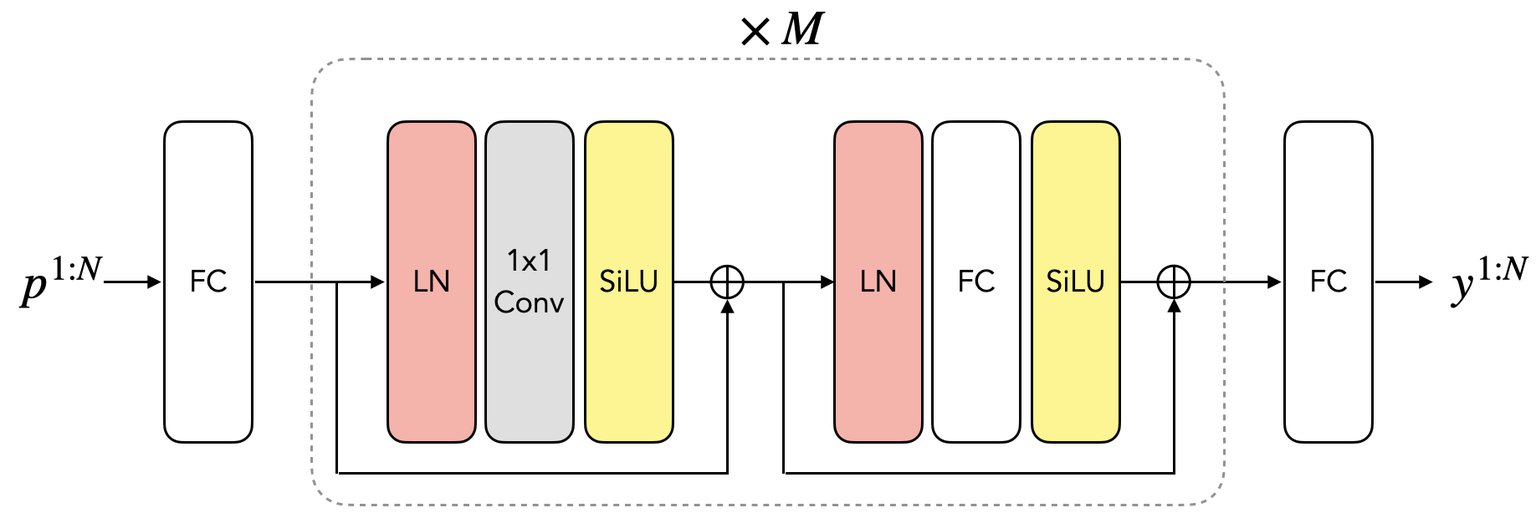}
  \caption{\textbf{The architecture of our MLP-based network.} 
  \textit{FC}, \textit{LN}, and \textit{SiLU} denote the fully connected layer, the layer normalization, and the SiLU activation layer respectively.
  \textit{1 $\times$ 1 Conv} denotes the 1D convolution layer with kernel size 1. Note that \textit{1 $\times$ 1 Conv} here is equivalent to a fully connected layer operating on the first dimension of the input tensor $\mathbb{R}^{N \times D}$, while the \textit{FC} layers operate on the last dimension. $N$ denotes the temporal dimension and $D$ denotes the dimension of the latent space. The middle block is repeated $M$ times. The first \textit{FC} layer projects input data to a latent space $\mathbb{R}^{N \times D}$ and the last one converts from latent space to the output space of full-body poses $\mathbb{R}^{N \times S}$. }
  \label{fig:mlp_network}
  \vspace{-3mm}
\end{figure}

\section{Related Work}

\subsection{Motion Tracking from Sparse Tracking Inputs}

The generation of full-body poses from sparse tracking signals of body joints has become an area of considerable interest within the research community. For instance, recent works such as~\cite{huang2018deep} have demonstrated the ability to track full bodies using only 6 IMU inputs and employing a bi-directional LSTM to predict SMPL body joints. Additionally, in~\cite{yang2021lobstr}, a similar approach is used to track with 4 IMU inputs, specifically the head, wrists, and pelvis. However, in the practical HMD setting, only 3 tracking signals are typically available: the head and 2 wrists. In this context, AvatarPoser~\cite{jiang2022avatarposer} provides a solution to the 3-point problem through the use of a transformer-based architecture. Other methods attempt to solve sparse input body tracking as a synthesis problem. To that extent, Aliakbarian \etal~\cite{aliakbarian2022flag} proposed a flow-based architecture derived from~\cite{dinh2016density}, while Dittadi \etal~\cite{dittadi2021full_VAE_HMD} opted for a Variational Autoencoder (VAE) method. While more complex methods have been developed that involve Reinforcement Learning, as seen in~\cite{winkler2022questsim,ye2022neural3points}, these approaches may struggle to simultaneously maintain accurate upper-body tracking while generating physically realistic motions.

In summary, all methods presented in this section either require more than three joints input or face difficulties in accurately predicting full body pose, particularly in the lower body region. Our proposed method, on the other hand, utilizes a custom diffusion model and employs a straightforward MLP-based architecture to predict full body pose with a high degree of accuracy, while utilizing only three IMU inputs.

\subsection{Diffusion Models and Motion Synthesis}

Diffusion models~\cite{sohl2015deep, ho2020denoising, nichol2021improved} are a class of likelihood-based generative models based on learning progressive noising and denoising of data. Diffusion models have recently have garnered significant attention in the field of image generation~\cite{dhariwal2021diffusion} due to their ability to significantly outperform popular GAN architectures~\cite{karras2019style, brock2018large} and is better suited for handling a large amount of data. Furthermore, diffusion models can support conditional generation, as evidenced by the classifier guidance approach presented in~\cite{dhariwal2021diffusion} and the CLIP-based text conditional synthesis for diffusion models proposed in~\cite{nichol2021glide}.

More recently, concurrent works have also extended diffusion models to motion synthesis, with particular focus on the text-to-motion task~\cite{zhang2022motiondiffuse, kim2022flame, tevet2022human}. However, these models are both complex in architecture and require multiple iterations at inference time. This hinders them unsuitable for real-time applications like VR body tracking. We circumvent this problem by designing a custom and efficient diffusion model. To the best of our knowledge, we present the first diffusion model solely purposed for solving motion reconstruction from sparse inputs. Our model leverages a simple MLP architecture, runs in real-time, and provides accurate pose predictions, particularly for lower bodies.

\subsection{Human Motion Synthesis}
\label{rw_human_motion}

Early works in human motion synthesis rose under the task of future motion prediction. Works around this task saw various modeling approaches ranging from sequence to sequence models~\cite{fragkiadaki2015recurrent} to graph modeling of each body part~\cite{Jain_2016_CVPR}. These supervised models were later replaced by generative methods~\cite{gui2018adversarial, li2018convolutional} based on Generative Adversarial Networks (GANs)~\cite{goodfellow2020generative}. Despite their leap forward, these approaches tend to diverge from realistic motion and require access to all body joint positions, making them impractical for avatar animation in VR~\cite{habibie2017recurrent}.

A second family of motion synthesis methods revolves around character control. In this setting, 
character motion must be generated according to user inputs and environmental constraints, such as the virtual environment properties. This research direction has practical applications in the field of computer gaming, where controller input is used to guide character motion. Taking inspiration from these constraints, Wang et al.~\cite{wang2019combining} formulated motion synthesis as a control problem by using a GAN architecture that takes direction and speed input into account. Similar efforts are found in~\cite{starke2020local}, where the method learns fast and dynamic character interactions that involve contacts between the body and other objects, given user input from a controller. These methods are impractical in a VR setting, where users want to drive motion using their real body pose instead of a controller.
\section{Method}
\subsection{Problem Formulation}
Our goal is to predict the whole body motion given sparse tracking signals, i.e.\ the orientation and translation of the headset and two hand controllers. To achieve this, we use a sequence of $N$ observed joint features $p^{1:N}=\{p^i\}^N_{i=1} \in \mathbb{R}^{N \times C}$ and aim to predict the corresponding whole-body poses $y^{1:N}=\{y^i\}^N_{i=1} \in \mathbb{R}^{N \times S}$ for each frame. The dimensions of the input/output joint features are represented by $C$ and $S$, respectively. We utilize the SMPL~\cite{loper2015smpl} model in this paper to represent human poses and follow the approach outlined in ~\cite{jiang2022avatarposer, dittadi2021full_VAE_HMD} to consider the first 22 joints of the SMPL model and disregard the joints on the hands and face. Thus, $y^{1:N}$ reflects the global orientation of the pelvis and the relative rotation of each joint. Following \cite{jiang2022avatarposer}, during inference, we initially pose the human model using the predicted rotations. Next, we calculate the character's global translation by accounting for the known head translation and subtracting the offset between the root joint and the head joint.

In the following section, we first introduce a simple MLP-based network for full-body motion synthesis based on sparse tracking signals. Then, we show how we further improve the performance by leveraging the proposed MLP-based architecture to power the conditional generative diffusion model, termed AGRoL.

\subsection{MLP-based Network}
\label{sec:mlp}
Our network architecture comprises only four types of components commonly employed in the realm of deep learning: fully connected layers (FC), SiLU activation layers~\cite{ramachandran2017swish}, 1D convolutional layers~\cite{lecun1989backpropagation} with kernel size 1 and an equal number of input and output channels, as well as layer normalization (LN)~\cite{ba2016layer}. It is worth noting that the 1D convolutional layer with a kernel size of 1 can also be interpreted as a fully connected layer operating along a different dimension. The details of our network architecture are demonstrated in Figure~\ref{fig:mlp_network}. Each block of the MLP network contains one convolutional and one fully connected layer, which is responsible for temporal and spatial information merging respectively. We use skip-connections as in ResNets~\cite{he2016deep} with Layer Norm~\cite{ba2016layernorm} as pre-normalization of the layers.  First, we project the input data $p^{1:N}$ to a higher dimensional latent space using a linear layer. And the last layer of the network projects from the latent space to the output space of full-body poses $y^{1:N}$.

\begin{figure}
  \centering
    \includegraphics[width=\linewidth]{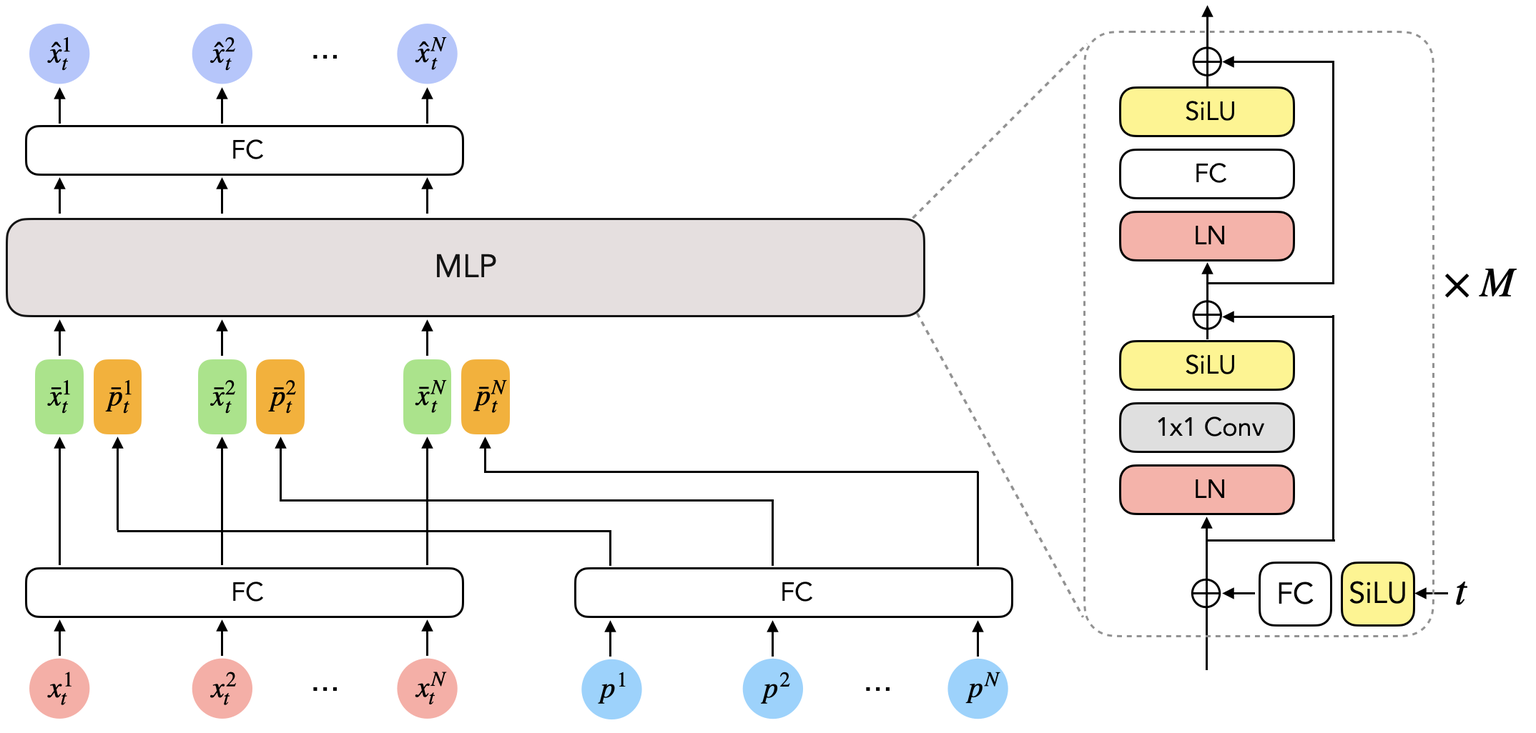}
  \caption{\textbf{The architecture of our MLP-based diffusion model.} $t$ is the noising step. $x_t^{1:N}$ denotes the motion sequence of length $N$ at step $t$, which is pure Gaussian noises when $t=T$. $p^{1:N}$ denotes the sparse upper body signals of length $N$. $\hat{x}_t^{1:N}$ denotes the denoised motion sequence at step $t$.}
  \label{fig:diffusion_pipeline}
  \vspace{-3mm}
\end{figure}

\subsection{Diffusion Model}\label{sec:diffusion_model}
Diffusion model~\cite{ho2020denoising, sohl2015deep} is a type of generative model which learns to reverse random Gaussian noise added by a Markov chain to recover desired data samples from the noise. In the forward diffusion process, given a sample motion sequence $x^{1:N}_0 \sim q(x^{1:N}_0)$ from the data distribution, the Markovian noising process can be written as:

\begin{equation}
    q(x^{1:N}_t|x^{1:N}_{t-1}) := \mathcal{N}(x^{1:N}_t; \sqrt{\alpha_t}x^{1:N}_{t-1}, (1-\alpha_t)I),
\end{equation}
where $\alpha_t \in (0,1)$ is constant hyper-parameter and $I$ is the identity matrix. $x_T^{1:N}$ tends to an isotropic Gaussian distribution when $T \to \infty$. Then, in the reverse diffusion process, a model $p_{\theta}$ with parameters $\theta$ is trained to generate samples from input Gaussian noise $x_T \sim \mathcal{N}(0,I)$ with variance $\sigma_t^2$ that follows a fixed schedule. Formally,

\begin{equation}
    p_{\theta}(x^{1:N}_{t-1}|x^{1:N}_t) := \mathcal{N}(x^{1:N}_{t-1}; \mu_{\theta}(x_t,t), \sigma_t^2 I),
\end{equation}

where $\mu_{\theta}$ could be reformulated~\cite{ho2020denoising} as

\begin{equation}
    \mu_{\theta}(x_t, t) = \frac{1}{\sqrt{\alpha_t}} ( x_t - \frac{1 - \alpha_t}{\sqrt{1 - \bar{\alpha}_t}} \epsilon_{\theta}(x_t, t)),
\end{equation}
where $\bar{\alpha}_t = \alpha_1 \cdot \alpha_2 ... \cdot \alpha_t$. So the model has to learn to predict noise $\epsilon_{\theta}(x_t, t)$ from $x_t$ and timestep $t$. 

In our case, we want to use the diffusion model to generate sequences of full-body poses conditioned on the sparse tracking of joint features $p^{1:N}$. Thus, the reverse diffusion process becomes conditional: $p_{\theta}(x^{1:N}_{t-1}|x^{1:N}_t, p^{1:N})$. Moreover, we follow ~\cite{ramesh2022hierarchical} to directly predict the clean body poses $x_0^{1:N}$ instead of predicting the residual noise $\epsilon_{\theta}(x_t, t)$. The objective function is then formulated as
\begin{equation}
\label{equat:objective_function}
    \mathcal{L}_{dm} = \mathbb{E}_{x^{1:N}_0 \sim q(x^{1:N}_0), t} \left[ \parallel x^{1:N}_0 - \hat{x}_0^{1:N} \parallel^2_2 \right]
\end{equation}
where the $\hat{x}_0^{1:N} = f_{\theta}(x^{1:N}, p^{1:N}, t)$ denotes the output of our model $f_{\theta}$.

We use the MLP architecture proposed in Sect.~\ref{sec:mlp} as the backbone for the model $f_{\theta}$ that predicts the full-body poses. At time step $t$, the motion features $x^{1:N}_t$ and the observed joints feature $p^{1:N}$ are first passed separately through a fully connected layer to obtain the latent features $\bar{x}^{1:N}_t$ and $\bar{p}^{1:N}$: 

\begin{equation}
    \bar{x}^{1:N}_t = \text{FC}_0(x^{1:N}_t),
\end{equation}
\begin{equation}
    \bar{p}^{1:N} = \text{FC}_1(p^{1:N}).
\end{equation}

Then these features are concatenated together and fed to the MLP backbone: 
    $\hat{x}_0^{1:N} = \text{MLP}(\text{Concat}(\bar{x}^{1:N}_t, \bar{p}^{1:N}), t)$.

\noindent\textbf{Block-wise Timestep Embedding.} 
When utilizing diffusion models, the embedding of the timestep $t$ is often included as an additional input to the network. To achieve this, a common approach is to concatenate the timestep embedding with the input, similar to positional embedding used in transformer-based methods~\cite{vaswani2017attention, dosovitskiy2020image}. However, since our network is mainly composed of FC layers, that mix the input features indiscriminately~\cite{tolstikhin2021mlp}, the time step embedding information can easily be lost after several layers, which hinders learning the denoising process and results in predicted motions with severe jittering artifacts, as shown in Section~\ref{sec:ablation_time_embedding}. In order to address the issue of losing time step embedding information in our network, we introduce a novel strategy that repetitively injects the time step embedding into every block of the MLP network. This process involves projecting the timestep embedding to match the input feature dimensions through a fully connected layer and a SiLU activation layer. The details of our pipeline are shown in Figure~\ref{fig:diffusion_pipeline}. Unlike previous work, such as~\cite{ho2020denoising}, which predicts a scale and shift factor for each block from the timestep embedding,our proposed approach directly adds the timestep embedding projections to the input activations of each block. Our experiments in Sect.~\ref{sec:experiments} validate that this approach significantly reduces jittering issues and enables the synthesis of smooth motions.
\section{Experiments}\label{sec:experiments}

\begin{table*}[t!]
    \centering
    \footnotesize
    \begin{tabular}{l|cccccccccc}
        \toprule
         Method & MPJRE & MPJPE & MPJVE & Hand PE & Upper PE & Lower PE & Root PE & Jitter & Upper Jitter & Lower Jitter \\
        \midrule
         Final IK & 16.77 & 18.09 & 59.24 & - & - & - & - & - & - & -\\
         LoBSTr & 10.69 & 9.02 & 44.97 & - & - & - & - & - & - & -\\
         VAE-HMD & 4.11 & 6.83 & 37.99 & - & - & - & - & - & - & -\\
         AvatarPoser*  & 3.08 & 4.18 & 27.70 & \underline{2.12} & \underline{1.81} & 7.59 & \bf 3.34 & 14.49 & \underline{7.36} & 24.81 \\
         MLP (Ours) & \underline{2.69} & \underline{3.93} & \underline{22.85} & 2.62 & 1.89 & \underline{6.88} & \underline{3.35} & \underline{13.01} & 9.13 & \underline{18.61} \\
         \rowcolor{gray!9} \textbf{AGRoL (Ours)} & \bf 2.66 & \bf 3.71 & \bf 18.59 & \bf 1.31 & \bf 1.55 & \bf 6.84 & 3.36 & \bf 7.26 & \bf 5.88 & \bf 9.27 \\
         \midrule
         GT & 0 & 0 & 0 & 0 & 0 & 0 & 0 & 4.00 & 3.65 & 4.52 \\
        \bottomrule
    \end{tabular}
    \caption{Comparison of our approach with state-of-the-art methods on a subset of AMASS dataset following~\cite{jiang2022avatarposer}. We report \textit{MPJPE} [cm], \textit{MPJRE} [$\deg$], \textit{MPJVE} [cm$/$s], Jitter [$10^2$m$/\text{s}^3$] metrics. 
    AGRoL achieves the best performance on \textit{MPJPE}, \textit{MPJRE} and \textit{MPJVE}, and outperforms other models, especially on the \textit{Lower PE} (Lower body Position Error) and \textit{Jitter} metrics, which shows that our model generates accurate lower body movement and smooth motions.
    }
    \label{tab:benchmark}
\end{table*}

Our models are trained and evaluated on the AMASS dataset~\cite{AMASS:ICCV:2019}. To compare with previous methods, we use two different settings for training and testing. In the first setting, we follow the approach of~\cite{jiang2022avatarposer}, which utilizes three subsets of AMASS: CMU~\cite{cmuWEB}, BMLr~\cite{BMLrub}, and HDM05~\cite{MPI_HDM05}. In the second setting, we adopt the data split employed in several recent works, including~\cite{dittadi2021full_VAE_HMD, aliakbarian2022flag, rempe2021humor}. This approach employs a larger set of training data, including CMU~\cite{cmuWEB}, MPI Limits~\cite{akhter2015pose}, Total Capture~\cite{trumble2017total}, Eyes Japn~\cite{eyesJapan}, KIT~\cite{mandery2015kit}, BioMotionLab~\cite{BMLrub}, BMLMovi~\cite{ghorbani2020movi}, EKUT~\cite{mandery2015kit}, ACCAD~\cite{ACCAD}, MPI Mosh~\cite{loper2014mosh}, SFU~\cite{SFU}, and HDM05~\cite{MPI_HDM05} as training data, while HumanEval~\cite{sigal2010humaneva} and Transition~\cite{AMASS:ICCV:2019} serve as testing data.

In both settings, we adopt the SMPL~\cite{loper2015smpl} human model for the human pose representation and train our model to predict the global orientation of the root joint and relative rotation of the other joints.

\begin{table}[t!]
    \centering
    \resizebox{0.97\linewidth}{!}{%
    \begin{tabular}{l|ccccccccc}
        \toprule
         Method & MPJRE & MPJPE & MPJVE & Jitter\\
        \midrule
         VAE-HMD$^\dagger$~\cite{dittadi2021full_VAE_HMD} & - & 7.45 & - & - \\
         HUMOR$^\dagger$~\cite{rempe2021humor} & - & 5.50 & - & - \\
         FLAG$^\dagger$~\cite{aliakbarian2022flag} & - & 4.96 & - & - \\
         \midrule
         \midrule
         AvatarPoser* & 4.70 & \underline{6.38} & 34.05 & \underline{10.21} \\
         MLP (Ours) & \underline{4.33} & 6.66 & \underline{33.58} & 21.74 \\
         \rowcolor{gray!9} 
         AGRoL (Ours) & \textbf{4.30} & \textbf{6.17} & \textbf{24.40} & \textbf{8.32} \\
         \midrule
         GT & 0 & 0 & 0 & 2.93 \\
        \bottomrule
    \end{tabular}
    }
    \caption{Comparison of our approach with state-of-the-art methods on AMASS dataset following the protocol of~\cite{dittadi2021full_VAE_HMD, rempe2021humor, aliakbarian2022flag}. We report the MPJPE [cm], MPJRE [$\deg$], MPJVE [cm$/$s], and Jitter [$10^2$m$/\text{s}^3$] metrics. The * denotes that we retrained the AvatarPoser using public code. $\dagger$ denotes methods that use pelvis location and rotation during inference, which are not directly comparable to our method, as we assume that the pelvis information is not available during the training and the testing. The best results are in bold, and the second-best results are underlined.
    }
    \label{tab:benchmark_full_amass}
\end{table}

\subsection{Implementation Details}
We represent the joint rotations by the 6D reparametrization~\cite{zhou2019continuity} due to its simplicity and continuity. Thus, for the sequences of body poses $y^{1:N} \in \mathbb{R}^{N \times S}$, $S = 22 \times 6$. %
The observed joint features $p^{1:N} \in \mathbb{R}^{N \times C}$ consists of the orientation, translation, orientation velocity and translation velocity of the head and hands in global coordinate system. Additionally, we adopt 6D reparametrization for the orientation and orientation velocity, thus $C = 18 \times 3$. Unless otherwise stated, we set the frame number $N$ to 196.

\vspace{-3mm}
\paragraph{MLP Network} We build our MLP network using 12 blocks ($M  = 12$). All latent features in the MLP network have the same shape of $N \times 512$. The network is trained with batch size 256 and Adam optimizer~\cite{kingma2014adam}. The learning rate is set to 3e-4 at the beginning and drops to 1e-5 after 200000 iterations. The weight decay is set to 1e-4 for the entire training. During inference, we apply our model in an auto-regressive manner for the longer sequences.

\vspace{-3mm}
\paragraph{MLP-based Diffusion Model (AGRoL)} We keep the MLP network architecture unchanged in the diffusion model. To inject the time step embedding used in the diffusion process in the network, in each MLP block, we pass the time step embedding to a fully connected layer and a SiLU activation layer~\cite{ramachandran2017swish} and sum it with the input feature. The network is trained with exactly the same hyperparameters as the MLP network, with the exception of using the AdamW~\cite{loshchilov2017decoupled} as optimizer. During training, we set the sampling step to 1000 and employ a cosine noise schedule~\cite{nichol2021improved}. However, to expedite the inference speed, we leverage the DDIM~\cite{song2020denoising} technique, which allows us to sample only 5 steps instead of 1000 during inference.

All experiments were carried out on a single NVIDIA V100 graphics card, using the PyTorch framework~\cite{paszke2019pytorch}.

\subsection{Evaluation Metrics}

In line with previous works~\cite{jiang2022avatarposer, yi2022physical, dittadi2021full_VAE_HMD, rempe2021humor}, we adopt nine evaluation metrics that we group into three categories.

    \textbf{Rotation-related metric}:  Mean Per Joint Rotation Error [degrees] (\textit{MPJRE}) measures the average relative rotation error for all joints.
    
    \textbf{Velocity-related metrics}: These include Mean Per Joint Velocity Error [cm/s] (\textit{MPJVE}) and \textit{Jitter}. \textit{MPJVE} measures the average velocity error for all joints, while \textit{Jitter}~\cite{yi2022physical} evaluates the mean jerk (change in acceleration over time) of all body joints in global space, expressed in $10^2\text{m/s}^3$. \textit{Jitter} is an indicator of motion smoothness.
    
    \textbf{Position-related metrics}. Mean Per Joint Position Error [cm] (\textit{MPJPE}) quantifies the average position error across all joints. \textit{Root PE} assesses the position error of the root joint, whereas \textit{Hand PE} calculates the average position error for both hands. \textit{Upper PE} and \textit{Lower PE} estimate the average position error for joints in the upper and lower body, respectively.

\begin{figure*}[t!]
    \centering
    \begin{subfigure}[b]{0.33\linewidth}
    \includegraphics[width=1.0\linewidth,height=0.5\linewidth]{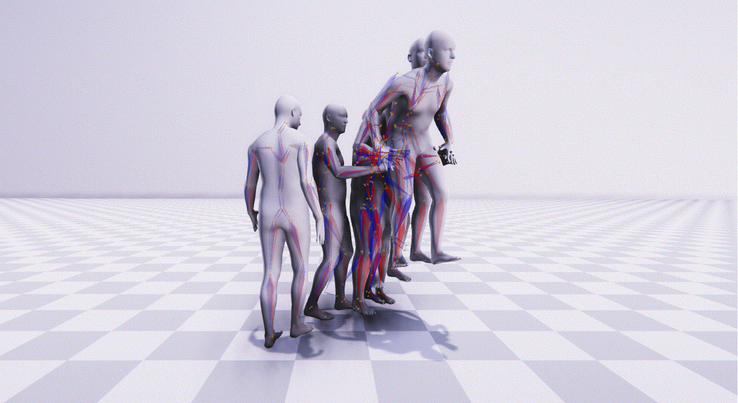}
    \end{subfigure}
    \begin{subfigure}[b]{0.33\linewidth}
    \includegraphics[width=1.0\linewidth,height=0.5\linewidth]{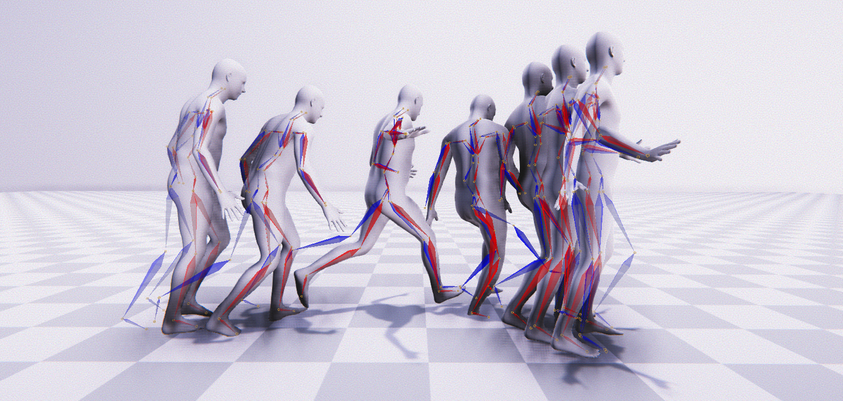}
    \end{subfigure}
    \begin{subfigure}[b]{0.33\linewidth}
    \includegraphics[width=1.0\linewidth,height=0.5\linewidth]{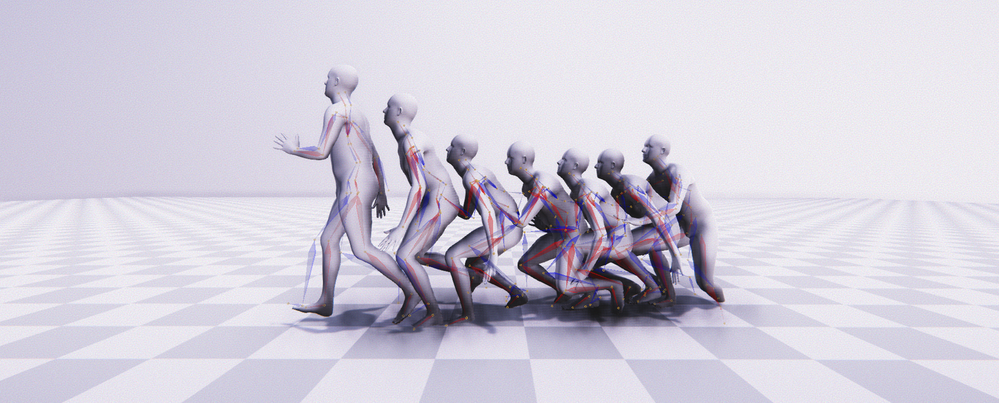}
    \end{subfigure}
    \begin{subfigure}[b]{0.33\linewidth}
    \includegraphics[width=1.0\linewidth,height=0.5\linewidth]{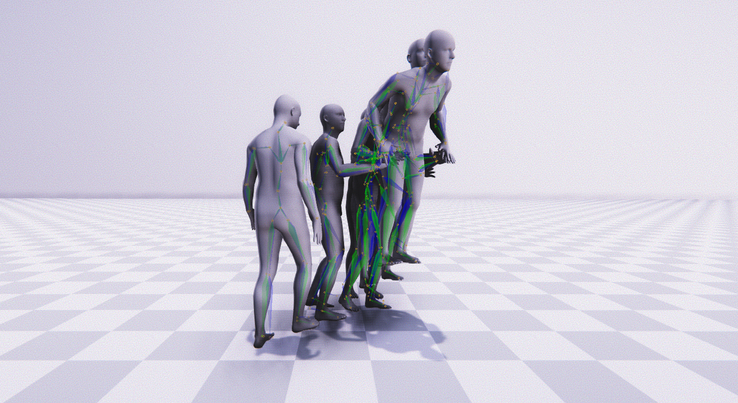}
    \end{subfigure}
    \begin{subfigure}[b]{0.33\linewidth}
    \includegraphics[width=1.0\linewidth,height=0.5\linewidth]{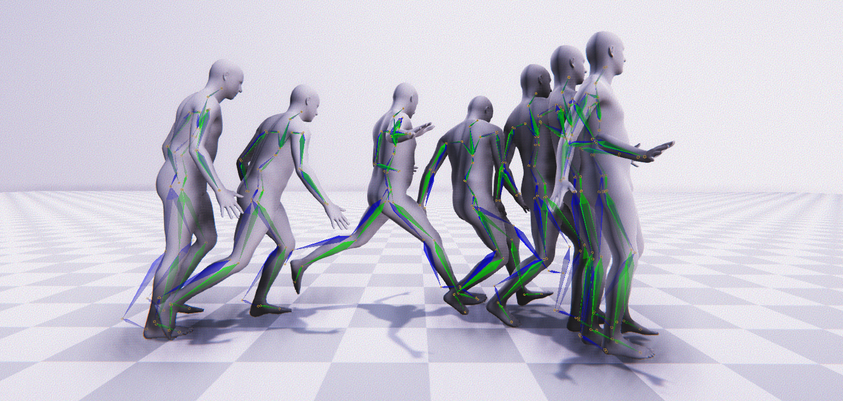}
    \end{subfigure}
    \begin{subfigure}[b]{0.33\linewidth}
    \includegraphics[width=1.0\linewidth,height=0.5\linewidth]{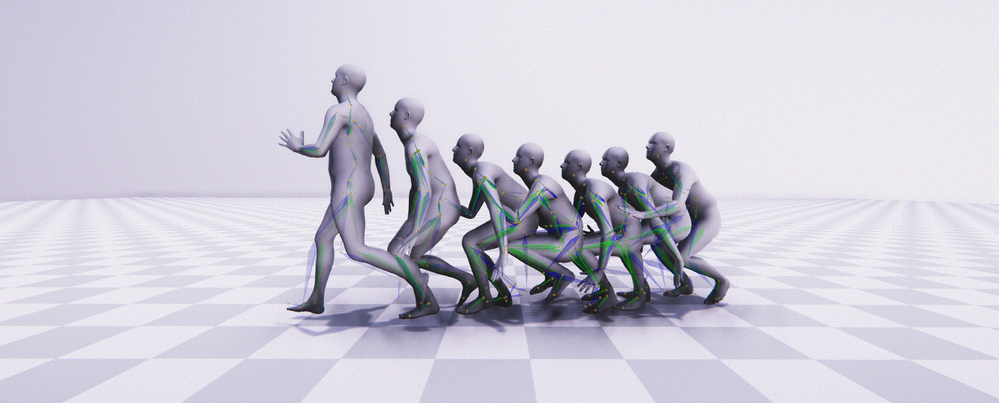}
    \end{subfigure}
    \vspace{-3mm}
    \caption{Qualitative comparison between AGRoL (top) and AvatarPoser~\cite{jiang2022avatarposer} (bottom) on test sequences from AMASS dataset. We visualize the predicted skeletons and render human body meshes. \textbf{Top:} AvatarPoser predictions in \textcolor{traj_red}{red}. \textbf{Bottom:} AGRoL predictions in \textcolor{traj_green}{green}. In both rows, the \textcolor{blue}{blue} skeletons denote the ground truth motion. We observe that motions predicted by AGRoL are closer to ground truth compared to the predictions of AvatarPoser.}
    \label{fig:reconstruction_error}
\end{figure*}

\begin{table*}[t!]
    \centering
    \footnotesize
    \begin{tabular}{l|ccccccccc}
        \toprule
         Method & \#Params & MPJRE & MPJPE & MPJVE & Hand PE & Upper PE & Lower PE & Root PE & Jitter\\
        \midrule
         AGRoL-AvatarPoser & 2.89M & 4.31 & 6.71 & 27.65 & 1.47 & 2.56 & 12.69 & 6.69 & 9.57 \\
         AGRoL-AvatarPoser-Large & 7.63M & \underline{2.86} & \underline{4.04} & 21.90 & \bf 1.29 & \underline{1.62} & \underline{7.53} & \underline{3.64} & 9.94 \\
         AGRoL-Transformer & 7.03M & 3.01 & 4.41 & \underline{20.33} & 2.97 & 2.13 & 7.71 & 3.88 & \bf 6.45 \\
         \rowcolor{gray!9} \textbf{AGRoL (Ours)} & 7.48M & \bf 2.66 & \bf 3.71 & \bf 18.59 & \underline{1.31} & \bf 1.55 & \bf 6.84 & \bf 3.36 & \underline{7.26} \\
        \bottomrule
    \end{tabular}
    \caption{Ablation study of network architectures in our diffusion model. We replace the proposed MLP backbone with other architectures and train several versions of the diffusion model with the same hyperparameters. The \textit{AvatarPoser-Large} denotes the backbone with the same architecture as AvatarPoser~\cite{jiang2022avatarposer} but with more transformer layers. \textit{AGRoL-Transformer} is the AGRoL version with the transformer backbone from \cite{tevet2022human}. The \textit{AGRol (ours)} with our MLP backbone outperforms all other backbones on most of the metrics.}
    \label{tab:ablation_arch}
\end{table*}

\subsection{Evaluation Results}
\begin{figure*}[t!]
    \centering
    \begin{subfigure}[b]{0.33\linewidth}
    \caption{GT}
    \includegraphics[width=1.0\linewidth]{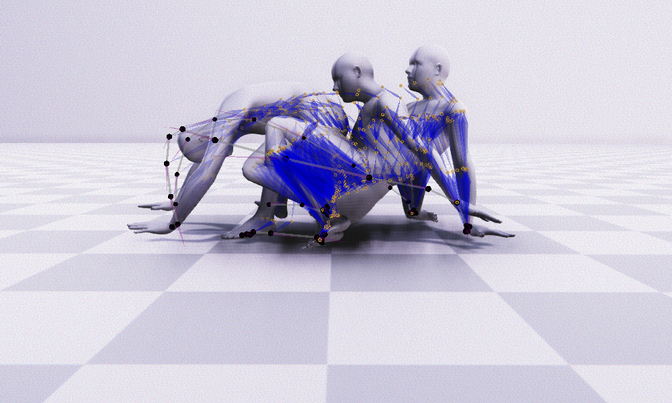}
    \end{subfigure}
    \begin{subfigure}[b]{0.33\linewidth}
    \caption{AGRoL (Ours)}
    \includegraphics[width=1.0\linewidth]{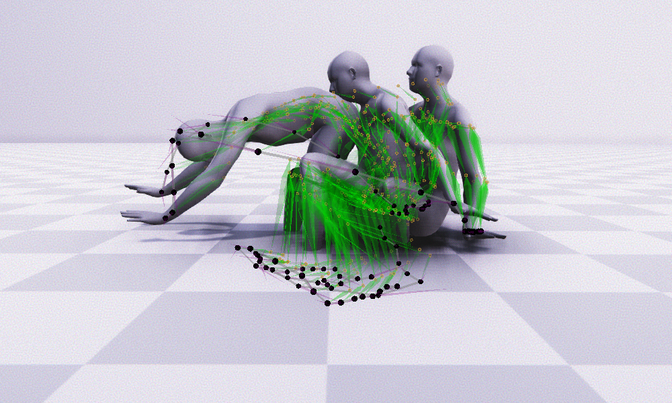}
    \end{subfigure}
    \begin{subfigure}[b]{0.33\linewidth}
    \caption{AvatarPoser\cite{jiang2022avatarposer}}
    \includegraphics[width=1.0\linewidth]{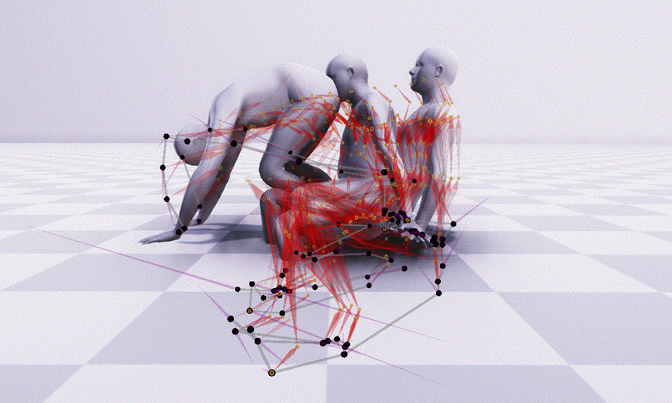}
    \end{subfigure}
    \begin{subfigure}[b]{0.33\linewidth}
    \includegraphics[width=1.0\linewidth]{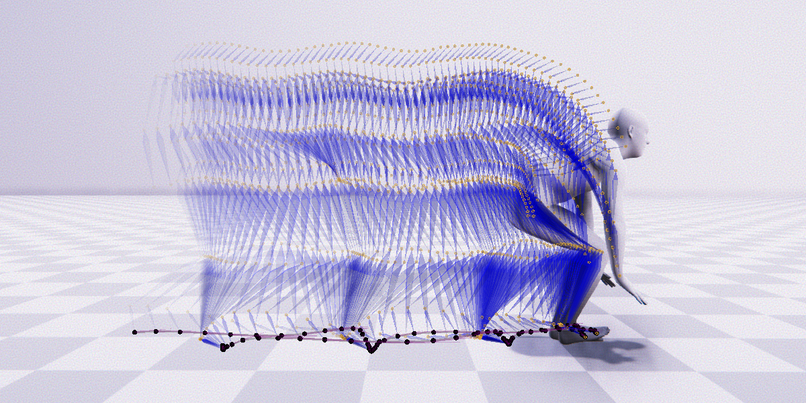}
    \end{subfigure}
    \begin{subfigure}[b]{0.33\linewidth}
    \includegraphics[width=1.0\linewidth]{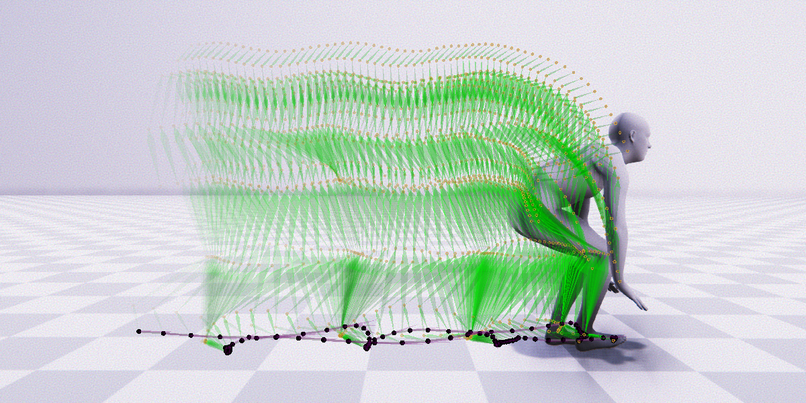}
    \end{subfigure}
    \begin{subfigure}[b]{0.33\linewidth}
    \includegraphics[width=1.0\linewidth]{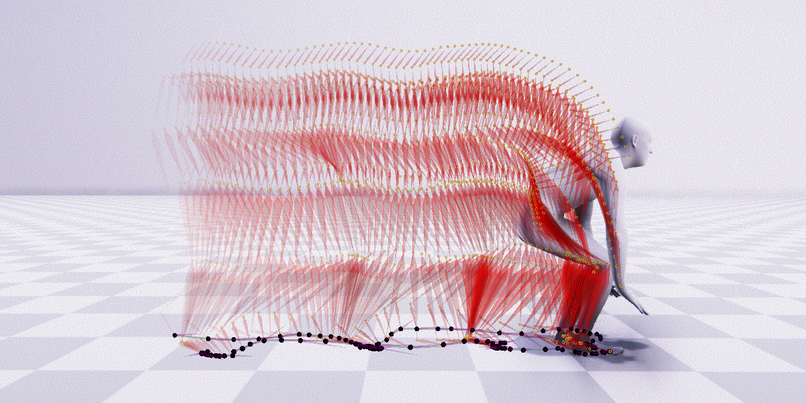}
    \end{subfigure}
    \caption{Motion trajectory visualization for predicted motions. \textbf{(a)} The ground truth motion with \textcolor{traj_blue}{blue} skeletons; \textbf{(b)} motion predicted by AGRoL with \textcolor{traj_green}{green} skeletons; \textbf{(c)} motion predicted by AvatarPoser with \textcolor{traj_red}{red} skeletons. The \textcolor{traj_bright_purple}{purple} vectors denote the velocity vectors of the corresponding joints. Observing the motion trajectories, we can see jittering and foot sliding issues more clearly. Smooth motion typically exhibits regular pose trajectories with the velocity vector of each joint changing steadily. The density of joint trajectories varies with walking speed; trajectories become denser as the individual slows down. Therefore, in the absence of foot sliding, we should observe a significantly high density of points when a foot makes contact with the ground. The black dots in the bottom row represent the trajectories of the foot joints. We notice more pronounced spikes in the density of foot trajectories for AGRoL compared to AvatarPoser.}
    \label{fig:trajectory}
\end{figure*}

We evaluate our method on the AMASS dataset with two different protocols. As shown in Table~\ref{tab:benchmark} and Table~\ref{tab:benchmark_full_amass}, our MLP network can already surpass most of the previous methods and achieves comparable results with the state-of-the-art method~\cite{jiang2022avatarposer}, demonstrating the effectiveness of the proposed simple MLP architecture. 
By leveraging the diffusion process and the proposed MLP backbone, the AGRoL model remarkably boosts the performance of the MLP network, surpassing all previous methods in all metrics\footnote{Except for insignificant 0.2 mm difference in Root PE in Tab.~\ref{tab:benchmark}.}. Moreover, our proposed AGRoL model significantly improves the smoothness of the generated motion, as reflected by the reduced \textit{Jitter} error compared to other methods. We visualize some examples in Figure~\ref{fig:reconstruction_error} and Figure~\ref{fig:trajectory}. In Figure~\ref{fig:reconstruction_error} we show the comparison of the reconstruction error between AGRoL and AvatarPoser. In Figure~\ref{fig:trajectory}, by visualizing the pose trajectories, we demonstrate the comparison of the smoothness and foot contact quality between AGRoL and AvatarPoser.\footnote{Please visit \href{https://dulucas.github.io/agrol/}{https://dulucas.github.io/agrol} for more visual examples.}

\subsection{Ablation Studies}
In this section, we ablate our methods on AMASS dataset. We first compare our proposed MLP architecture with other backbones in the context of the diffusion model in Section~\ref{sec:ablation_arch} to highlight the superiority of our MLP network. Then we investigate the importance of time step embedding for our diffusion model and evaluate different strategies for adding the time step embedding in Section~\ref{sec:ablation_time_embedding}. Finally, we analyze the impact of  %
the number of sampling steps used during inference in Section~\ref{sec:ablation_sampling_step}.

\subsubsection{Architecture}
\label{sec:ablation_arch}
To validate the effectiveness of our proposed MLP backbone in the diffusion model setup, we conduct experiments where we replace our MLP network in AGRoL with other types of backbones and compare them. Specifically, we consider two alternative backbone architectures: the network from AvatarPoser~\cite{jiang2022avatarposer} and the transformer network from Tevet et  al.~\cite{tevet2022human}. In transformer networks, instead of repetitively injecting the time positional embedding to every block, we concatenate the time positional embedding with the input features $\bar{x}^{1:N}$ and $\bar{p}^{1:N}$ before being fed to transformer layers. We apply the same technique to the AvatarPoser backbone as this model is also based on transformer layers. 
To ensure a fair comparison, we train AGRoL with two versions of AvatarPoser backbone. The first one, AGRoL-AvatarPoser, uses the architecture that follows exactly the same settings as described in the original paper~\cite{jiang2022avatarposer}, while the second one, AGRoL-AvatarPoser-Large, incorporates additional transformer layers to achieve a comparable size to our AGRoL model with MLP backbone. Similarly, we increase the number of layers in the transformer backbone~\cite{tevet2022human} and train AGRoL-Transformer.
As shown in Tab.~\ref{tab:ablation_arch}, the AGRoL diffusion model with the proposed MLP backbone achieves superior results compared to the versions with other backbones.

\subsubsection{Diffusion Time Step Embedding}
\label{sec:ablation_time_embedding}
In this section, we study the importance of time step embedding. Time step embedding is often used in diffusion-based models~\cite{dhariwal2021diffusion, zhang2022motiondiffuse} to indicate the noise level $t$ during the diffusion process. We use the sinusoidal positional embedding~\cite{vaswani2017attention} as the time step embedding. 
Although the AGRoL without time step embedding (see Tab.~\ref{tab:ablation_time_embedding}) can still attain reasonable performance on metrics related to position errors and rotation errors, the performance on metrics related to velocity errors (\textit{MPJVE} and \textit{Jitter}) is severely degraded. 
This outcome is expected as the absence of the time step embedding implies that the model is not aware of the current denoising step, rendering it unable to denoise accurately.

We now ablate three strategies for utilizing the time step embedding in our network: \textit{Add}, \textit{Concat}, and \textit{RepIn}. \textit{RepIn} (Repetitive Injection) repetitively passes the time step embedding through a linear layer and injects the results into every block of the MLP network. In contrast, \textit{Add} and \textit{Concat} inject the time step embedding only once at the beginning of the network. 

Before inputting it into the network, the time step embedding is passed through a fully connected layer and a SiLU activation to obtain a latent feature $u_t \in \mathbb{R}^{1 \times D}$. 
\textit{Add} sums the $u_t$ with the input features $\bar{x}^{1:N}_t$ and $\bar{p}^{1:N}$, the output of the network then becomes $\hat{x}^{1:N}_0 = \text{MLP}(\text{Concat}(\bar{x}^{1:N}_t, \bar{p}^{1:N}) + u_t)$, where vector $u_t$ is broadcasted along the first dimension.
\textit{Concat} concatenates the $u_t$ with the input features $\bar{x}^{1:N}_t$ and $\bar{p}^{1:N}$, resulting in $\hat{x}^{1:N}_0 = \text{MLP}(\text{Concat}(\bar{x}^{1:N}_t, \bar{p}^{1:N}, u_t))$. 

\textit{RepIn} is our proposed strategy for injecting the time step embedding. Specifically, for each block of the MLP network, we project the time step embedding separately using a fully connected layer and a SiLU activation, then we add the obtained features $u_{t,j}$ to the input features of the correspondent block, where $j \in [0,..M]$ and $M$ is the number of blocks. As shown in Table~\ref{tab:ablation_time_embedding}, our proposed strategy can largely improve the velocity-related metrics and alleviate the jittering issues and generate smooth motion.

\begin{table*}[t!]
    \centering
    \footnotesize
    \begin{tabular}{l|cccccccc}
        \toprule
         Method & MPJRE & MPJPE & MPJVE & Hand PE & Upper PE & Lower PE & Root PE & Jitter\\
        \midrule
         w/o Time & \underline{2.68} & \bf 3.63 & 22.80 & 1.36 & \bf 1.54 & \bf 6.67 & \bf 3.25 & 15.23 \\
         Add & 2.80 & 4.01 & 23.60 & 1.40 & 1.64 & 7.44 & 3.59 & 15.02 \\
         Concat & 2.72 & 3.79 & 21.99 & 1.31 & 1.57 & 7.00 & 3.43 & 13.30 \\
         \rowcolor{gray!9} RepIn (Ours) & \bf 2.66 & \underline{3.71} & \bf 18.59 & \bf 1.31 & \underline{1.55} & \underline{6.84} & \underline{3.36} & \bf 7.26 \\
        \bottomrule
    \end{tabular}
    \caption{Ablation of the time step embedding. \textit{w/o Time} denotes the results of AGRoL without time step embedding. \textit{Add} sums up the features from time step embedding with the input features. \textit{Concat} concatenates the features from time step embedding with the input features. In \textit{Add} and \textit{Concat}, the time step embedding is only fed once at the top of the network. \textit{RepIn} (Repetitive Injection) denotes our strategy to inject the time step embedding into every block of the network. The time step embedding mainly affects the \textit{MPJVE} and \textit{Jitter} metrics. Omiting the timestep embedding or adding it improperly results in high MPJVE and causes severe jittering issues.}
    \label{tab:ablation_time_embedding}
\end{table*}
\begin{table*}[t!]
    \centering
    \footnotesize
    \begin{tabular}{c|cccccccc}
        \toprule
         \# Sampling Steps & MPJRE & MPJPE & MPJVE & Hand PE & Upper PE & Lower PE & Root PE & Jitter\\
        \midrule
         2 & 3.17 & 4.93 & 20.03 & 2.19 & 2.12 & 8.98 & 4.61 & \bf 6.90 \\
         \rowcolor{gray!9} 5 & \bf 2.66 & \underline{3.71} & \bf 18.59 & \bf 1.31 & \bf 1.55 & \underline{6.84} & \underline{3.36} & \underline{7.26} \\
         10 & \underline{2.68} & \bf 3.69 & \underline{19.55} & \underline{1.39} & \bf 1.55 & \bf 6.77 & \bf 3.31 & 7.51 \\
         100 & 2.84 & 3.93 & 23.50 & 1.62 & 1.67 & 7.19 & 3.51 & 9.64 \\
         1000 & 2.97 & 4.14 & 27.25 & 1.82 & 1.78 & 7.55 & 3.66 & 12.79
         \\
        \bottomrule
    \end{tabular}
    \vspace{-2mm}
    \caption{Ablation of the number of DDIM~\cite{song2020denoising} sampling steps during inference. The input and output length is fixed to $N=196$.}
    \label{tab:ablation_sampling_step}
\end{table*}
\begin{table*}[t!]
    \centering
    \footnotesize
    \begin{tabular}{l|cccccccc}
        \toprule
         Methods & MPJRE & MPJPE & MPJVE & Hand PE & Upper PE & Lower PE & Root PE & Jitter\\
        \midrule
         AvatarPoser & 5.69 & 10.34 & 572.58 & 8.98 & 5.49 & 17.34 & 8.83 & 762.79 \\
         MLP & 5.37 & 10.76 & 107.82 & 12.43 & 6.48 & 16.94 & 8.74 & 92.51 \\
         Transformer & 4.44 & 8.62 & 135.99 & 7.29 & 5.28 & 13.44 & 10.32 & 147.09 \\
         \rowcolor{gray!9} AGRoL (Ours) & \textbf{4.20} & \textbf{6.38} & \textbf{96.85} & \textbf{5.27} & \textbf{3.86} & \textbf{10.03} & \textbf{6.67} & \textbf{33.35} \\
        \bottomrule
    \end{tabular}
    \caption{Robustness of the models to joints tracking loss. We evaluate the methods by randomly masking a portion (10\%) of input frames during the inference on AMASS dataset. We test each method 5 times and take the average results. AGRoL achieves the best performance among all the methods, which shows the robustness of our method against joint tracking loss.}
    \vspace{-4mm}
    \label{tab:ablation_robustness}
\end{table*}

\vspace{-1mm}
\subsubsection{Number of Sampling Steps during Inference}
\label{sec:ablation_sampling_step}
We ablate the number of sampling steps that we used during inference. In Tab.~\ref{tab:ablation_sampling_step}, we take the AGRoL model trained with $1000$ sampling steps and test it with a subset of diffusion steps during inference. We opted to use $5$ DDIM~\cite{song2020denoising} sampling steps as it enabled our model to achieve superior performance on most of the metrics while also being faster.

\subsection{Robustness to Tracking Loss}
\label{sec:robustness}
In this section, we study the robustness of our model against input joint tracking loss. 
In VR applications, it is common for the joint tracking signal to be lost on some frames when hands or controllers move out of the field of view, causing temporal discontinuity in the input signals.
We evaluate the performance of all available methods on tracking loss by randomly masking $10\%$ of input frames during inference, and present results in Tab.~\ref{tab:ablation_robustness}.
We observe that the performance of all previous methods is significantly degraded, indicating their lack of robustness against tracking loss. In comparison, AGRoL shows less degradation in accuracy, suggesting that our approach can accurately model motion even with highly sparse tracking inputs.

\subsection{Inference Speed}
\label{sec:inference_speed}
Our AGRoL model achieves real-time inference speed due to a lightweight architecture combined with DDIM sampling. A single AGRoL generation, that runs 5 DDIM sampling steps, produces 196 output frames in $35$~ms on a single NVIDIA V100 GPU.
Our predictive MLP model takes 196 frames as input and predicts a final result of 196 frames in a single forward pass. It is even faster and requires only $6$~ms on a single NVIDIA V100 GPU.

\section{Conclusion and Limitations}
In this paper, we presented a simple yet efficient MLP-based architecture with carefully designed building blocks which achieves competitive performance on the full-body motion synthesis task. Then we introduced AGRoL, a conditional diffusion model for full-body motion synthesis based on sparse tracking signal. AGRoL leverages a simple yet efficient conditioning scheme for structured human motion data. We demonstrated that our lightweight diffusion-based model generates realistic and smooth human motions while achieving real-time inference speed, making it suitable for online AR/VR applications. A notable limitation of our and related approaches is occasional floor penetration artifacts. Future work involves investigating this issue and integrating additional physical constraints into the model.

\section{Acknowledgements}
We thank Federica Bogo, Jonas Kohler, Wen Guo, Xi Shen for inspiring discussions and valuable feedback.

{\small
\bibliographystyle{ieee_fullname}
\bibliography{egbib}

\begin{thebibliography}{10}\itemsep=-1pt

\bibitem{ACCAD}
Osu accad.
\newblock \url{https://accad.osu.edu/research/motion-lab/system-data}.

\bibitem{SFU}
Sfu motion capture database.
\newblock \url{https://mocap.cs.sfu.ca/}.

\bibitem{akhter2015pose}
Ijaz Akhter and Michael~J Black.
\newblock Pose-conditioned joint angle limits for 3d human pose reconstruction.
\newblock In {\em Proceedings of the IEEE conference on computer vision and
  pattern recognition}, pages 1446--1455, 2015.

\bibitem{aliakbarian2022flag}
Sadegh Aliakbarian, Pashmina Cameron, Federica Bogo, Andrew Fitzgibbon, and
  Thomas~J Cashman.
\newblock Flag: Flow-based 3d avatar generation from sparse observations.
\newblock In {\em CVPR}, pages 13253--13262, 2022.

\bibitem{ba2016layer}
Jimmy~Lei Ba, Jamie~Ryan Kiros, and Geoffrey~E Hinton.
\newblock Layer normalization.
\newblock {\em arXiv preprint arXiv:1607.06450}, 2016.

\bibitem{ba2016layernorm}
Jimmy~Lei Ba, Jamie~Ryan Kiros, and Geoffrey~E Hinton.
\newblock Layer normalization.
\newblock {\em arXiv preprint arXiv:1607.06450}, 2016.

\bibitem{brock2018large}
Andrew Brock, Jeff Donahue, and Karen Simonyan.
\newblock Large scale gan training for high fidelity natural image synthesis.
\newblock {\em arXiv preprint arXiv:1809.11096}, 2018.

\bibitem{cmuWEB}
{Carnegie Mellon University}.
\newblock {CMU MoCap Dataset}.

\bibitem{dhariwal2021diffusion}
Prafulla Dhariwal and Alexander Nichol.
\newblock Diffusion models beat gans on image synthesis.
\newblock {\em NeurIPS}, 34:8780--8794, 2021.

\bibitem{dinh2016density}
Laurent Dinh, Jascha Sohl-Dickstein, and Samy Bengio.
\newblock Density estimation using real nvp.
\newblock {\em ICLR}, 2016.

\bibitem{dittadi2021full_VAE_HMD}
Andrea Dittadi, Sebastian Dziadzio, Darren Cosker, Ben Lundell, Thomas~J
  Cashman, and Jamie Shotton.
\newblock Full-body motion from a single head-mounted device: Generating smpl
  poses from partial observations.
\newblock In {\em ICCV}, pages 11687--11697, 2021.

\bibitem{dosovitskiy2020image}
Alexey Dosovitskiy, Lucas Beyer, Alexander Kolesnikov, Dirk Weissenborn,
  Xiaohua Zhai, Thomas Unterthiner, Mostafa Dehghani, Matthias Minderer, Georg
  Heigold, Sylvain Gelly, et~al.
\newblock An image is worth 16x16 words: Transformers for image recognition at
  scale.
\newblock {\em arXiv preprint arXiv:2010.11929}, 2020.

\bibitem{eyesJapan}
{Eyes, JAPAN Co. Ltd.}
\newblock {Eyes, Jappan}.

\bibitem{fragkiadaki2015recurrent}
Katerina Fragkiadaki, Sergey Levine, Panna Felsen, and Jitendra Malik.
\newblock Recurrent network models for human dynamics.
\newblock In {\em ICCV}, pages 4346--4354, 2015.

\bibitem{ghorbani2020movi}
Saeed Ghorbani, Kimia Mahdaviani, Anne Thaler, Konrad Kording, Douglas~James
  Cook, Gunnar Blohm, and Nikolaus~F Troje.
\newblock Movi: A large multipurpose motion and video dataset.
\newblock {\em arXiv preprint arXiv:2003.01888}, 2020.

\bibitem{goodfellow2020generative}
Ian Goodfellow, Jean Pouget-Abadie, Mehdi Mirza, Bing Xu, David Warde-Farley,
  Sherjil Ozair, Aaron Courville, and Yoshua Bengio.
\newblock Generative adversarial networks.
\newblock {\em Communications of the ACM}, 63(11):139--144, 2020.

\bibitem{gui2018adversarial}
Liang-Yan Gui, Yu-Xiong Wang, Xiaodan Liang, and Jos{\'e}~MF Moura.
\newblock Adversarial geometry-aware human motion prediction.
\newblock In {\em ECCV}, pages 786--803, 2018.

\bibitem{guo2022simple_mlp}
Wen Guo, Yuming Du, Xi Shen, Vincent Lepetit, Alameda-Pineda Xavier, and
  Moreno-Noguer Francesc.
\newblock Back to mlp: A simple baseline for human motion prediction.
\newblock In {\em Proceedings of the IEEE/CVF Winter Conference on Applications
  of Computer Vision (WACV)}, 2023.

\bibitem{habibie2017recurrent}
Ikhsanul Habibie, Daniel Holden, Jonathan Schwarz, Joe Yearsley, and Taku
  Komura.
\newblock A recurrent variational autoencoder for human motion synthesis.
\newblock In {\em British Machine Vision Conference}, 2017.

\bibitem{he2016deep}
Kaiming He, Xiangyu Zhang, Shaoqing Ren, and Jian Sun.
\newblock Deep residual learning for image recognition.
\newblock In {\em Proceedings of the IEEE conference on computer vision and
  pattern recognition}, pages 770--778, 2016.

\bibitem{ho2020denoising}
Jonathan Ho, Ajay Jain, and Pieter Abbeel.
\newblock Denoising diffusion probabilistic models.
\newblock {\em NeurIPS}, 33:6840--6851, 2020.

\bibitem{huang2018deep}
Yinghao Huang, Manuel Kaufmann, Emre Aksan, Michael~J Black, Otmar Hilliges,
  and Gerard Pons-Moll.
\newblock Deep inertial poser: Learning to reconstruct human pose from sparse
  inertial measurements in real time.
\newblock {\em ACM TOG}, 37(6):1--15, 2018.

\bibitem{Jain_2016_CVPR}
Ashesh Jain, Amir~R. Zamir, Silvio Savarese, and Ashutosh Saxena.
\newblock Structural-rnn: Deep learning on spatio-temporal graphs.
\newblock In {\em CVPR}, June 2016.

\bibitem{jiang2022avatarposer}
Jiaxi Jiang, Paul Streli, Huajian Qiu, Andreas Fender, Larissa Laich, Patrick
  Snape, and Christian Holz.
\newblock Avatarposer: Articulated full-body pose tracking from sparse motion
  sensing.
\newblock {\em ECCV}, 2022.

\bibitem{jiang2022transformer}
Yifeng Jiang, Yuting Ye, Deepak Gopinath, Jungdam Won, Alexander~W Winkler, and
  C~Karen Liu.
\newblock Transformer inertial poser: Attention-based real-time human motion
  reconstruction from sparse imus.
\newblock {\em arXiv preprint arXiv:2203.15720}, 2022.

\bibitem{karras2019style}
Tero Karras, Samuli Laine, and Timo Aila.
\newblock A style-based generator architecture for generative adversarial
  networks.
\newblock In {\em CVPR}, pages 4401--4410, 2019.

\bibitem{kaufmann2021pose}
Manuel Kaufmann, Yi Zhao, Chengcheng Tang, Lingling Tao, Christopher Twigg, Jie
  Song, Robert Wang, and Otmar Hilliges.
\newblock Em-pose: 3d human pose estimation from sparse electromagnetic
  trackers.
\newblock In {\em ICCV}, pages 11510--11520, 2021.

\bibitem{kim2022flame}
Jihoon Kim, Jiseob Kim, and Sungjoon Choi.
\newblock Flame: Free-form language-based motion synthesis \& editing.
\newblock {\em arXiv preprint arXiv:2209.00349}, 2022.

\bibitem{kingma2014adam}
Diederik~P Kingma and Jimmy Ba.
\newblock Adam: A method for stochastic optimization.
\newblock {\em arXiv preprint arXiv:1412.6980}, 2014.

\bibitem{lecun1989backpropagation}
Yann LeCun, Bernhard Boser, John~S Denker, Donnie Henderson, Richard~E Howard,
  Wayne Hubbard, and Lawrence~D Jackel.
\newblock Backpropagation applied to handwritten zip code recognition.
\newblock {\em Neural computation}, 1(4):541--551, 1989.

\bibitem{li2018convolutional}
Chen Li, Zhen Zhang, Wee~Sun Lee, and Gim~Hee Lee.
\newblock Convolutional sequence to sequence model for human dynamics.
\newblock In {\em CVPR}, pages 5226--5234, 2018.

\bibitem{li2021ai}
Ruilong Li, Shan Yang, David~A Ross, and Angjoo Kanazawa.
\newblock Ai choreographer: Music conditioned 3d dance generation with aist++.
\newblock In {\em Proceedings of the IEEE/CVF International Conference on
  Computer Vision}, pages 13401--13412, 2021.

\bibitem{loper2014mosh}
Matthew Loper, Naureen Mahmood, and Michael~J Black.
\newblock Mosh: Motion and shape capture from sparse markers.
\newblock {\em ACM Transactions on Graphics (ToG)}, 33(6):1--13, 2014.

\bibitem{loper2015smpl}
Matthew Loper, Naureen Mahmood, Javier Romero, Gerard Pons-Moll, and Michael~J
  Black.
\newblock Smpl: A skinned multi-person linear model.
\newblock {\em ACM TOG}, 34(6):1--16, 2015.

\bibitem{loshchilov2017decoupled}
Ilya Loshchilov and Frank Hutter.
\newblock Decoupled weight decay regularization.
\newblock {\em arXiv preprint arXiv:1711.05101}, 2017.

\bibitem{AMASS:ICCV:2019}
Naureen Mahmood, Nima Ghorbani, Nikolaus~F. Troje, Gerard Pons-Moll, and
  Michael~J. Black.
\newblock {AMASS}: Archive of motion capture as surface shapes.
\newblock In {\em ICCV}, pages 5442--5451, Oct. 2019.

\bibitem{mandery2015kit}
Christian Mandery, {\"O}mer Terlemez, Martin Do, Nikolaus Vahrenkamp, and Tamim
  Asfour.
\newblock The kit whole-body human motion database.
\newblock In {\em 2015 International Conference on Advanced Robotics (ICAR)},
  pages 329--336. IEEE, 2015.

\bibitem{MPI_HDM05}
M. M\"{u}ller, T. R\"{o}der, M. Clausen, B. Eberhardt, B. Kr\"{u}ger, and A.
  Weber.
\newblock Documentation mocap database {HDM05}.
\newblock Technical Report CG-2007-2, Universit\"{a}t Bonn, June 2007.

\bibitem{nichol2021glide}
Alex Nichol, Prafulla Dhariwal, Aditya Ramesh, Pranav Shyam, Pamela Mishkin,
  Bob McGrew, Ilya Sutskever, and Mark Chen.
\newblock Glide: Towards photorealistic image generation and editing with
  text-guided diffusion models.
\newblock {\em arXiv preprint arXiv:2112.10741}, 2021.

\bibitem{nichol2021improved}
Alexander~Quinn Nichol and Prafulla Dhariwal.
\newblock Improved denoising diffusion probabilistic models.
\newblock In {\em International Conference on Machine Learning}, pages
  8162--8171. PMLR, 2021.

\bibitem{paszke2019pytorch}
Adam Paszke, Sam Gross, Francisco Massa, Adam Lerer, James Bradbury, Gregory
  Chanan, Trevor Killeen, Zeming Lin, Natalia Gimelshein, Luca Antiga, et~al.
\newblock Pytorch: An imperative style, high-performance deep learning library.
\newblock {\em Advances in neural information processing systems}, 32, 2019.

\bibitem{petrovich2021action}
Mathis Petrovich, Michael~J Black, and G{\"u}l Varol.
\newblock Action-conditioned 3d human motion synthesis with transformer vae.
\newblock In {\em CVPR}, pages 10985--10995, 2021.

\bibitem{ramachandran2017swish}
Prajit Ramachandran, Barret Zoph, and Quoc~V Le.
\newblock Swish: a self-gated activation function.
\newblock {\em arXiv preprint arXiv:1710.05941}, 7(1):5, 2017.

\bibitem{ramesh2022hierarchical}
Aditya Ramesh, Prafulla Dhariwal, Alex Nichol, Casey Chu, and Mark Chen.
\newblock Hierarchical text-conditional image generation with clip latents.
\newblock {\em arXiv preprint arXiv:2204.06125}, 2022.

\bibitem{rempe2021humor}
Davis Rempe, Tolga Birdal, Aaron Hertzmann, Jimei Yang, Srinath Sridhar, and
  Leonidas~J Guibas.
\newblock Humor: 3d human motion model for robust pose estimation.
\newblock In {\em Proceedings of the IEEE/CVF International Conference on
  Computer Vision}, pages 11488--11499, 2021.

\bibitem{rezende2015variational}
Danilo Rezende and Shakir Mohamed.
\newblock Variational inference with normalizing flows.
\newblock In {\em International conference on machine learning}, pages
  1530--1538. PMLR, 2015.

\bibitem{shi2020motionet}
Mingyi Shi, Kfir Aberman, Andreas Aristidou, Taku Komura, Dani Lischinski,
  Daniel Cohen-Or, and Baoquan Chen.
\newblock Motionet: 3d human motion reconstruction from monocular video with
  skeleton consistency.
\newblock {\em ACM TOG}, 40(1):1--15, 2020.

\bibitem{sigal2010humaneva}
Leonid Sigal, Alexandru~O Balan, and Michael~J Black.
\newblock Humaneva: Synchronized video and motion capture dataset and baseline
  algorithm for evaluation of articulated human motion.
\newblock {\em International journal of computer vision}, 87(1):4--27, 2010.

\bibitem{sohl2015deep}
Jascha Sohl-Dickstein, Eric Weiss, Niru Maheswaranathan, and Surya Ganguli.
\newblock Deep unsupervised learning using nonequilibrium thermodynamics.
\newblock In {\em International Conference on Machine Learning}, pages
  2256--2265. PMLR, 2015.

\bibitem{song2020denoising}
Jiaming Song, Chenlin Meng, and Stefano Ermon.
\newblock Denoising diffusion implicit models.
\newblock {\em arXiv preprint arXiv:2010.02502}, 2020.

\bibitem{starke2020local}
Sebastian Starke, Yiwei Zhao, Taku Komura, and Kazi Zaman.
\newblock Local motion phases for learning multi-contact character movements.
\newblock {\em ACM TOG}, 39(4):54--1, 2020.

\bibitem{tevet2022human}
Guy Tevet, Sigal Raab, Brian Gordon, Yonatan Shafir, Amit~H Bermano, and Daniel
  Cohen-Or.
\newblock Human motion diffusion model.
\newblock {\em arXiv preprint arXiv:2209.14916}, 2022.

\bibitem{tolstikhin2021mlp}
Ilya~O Tolstikhin, Neil Houlsby, Alexander Kolesnikov, Lucas Beyer, Xiaohua
  Zhai, Thomas Unterthiner, Jessica Yung, Andreas Steiner, Daniel Keysers,
  Jakob Uszkoreit, et~al.
\newblock Mlp-mixer: An all-mlp architecture for vision.
\newblock {\em Advances in neural information processing systems},
  34:24261--24272, 2021.

\bibitem{BMLrub}
Nikolaus~F. Troje.
\newblock Decomposing biological motion: {A} framework for analysis and
  synthesis of human gait patterns.
\newblock {\em Journal of Vision}, 2(5):2--2, Sept. 2002.

\bibitem{trumble2017total}
Matthew Trumble, Andrew Gilbert, Charles Malleson, Adrian Hilton, and John
  Collomosse.
\newblock Total capture: 3d human pose estimation fusing video and inertial
  sensors.
\newblock In {\em Proceedings of 28th British Machine Vision Conference}, pages
  1--13, 2017.

\bibitem{vaswani2017attention}
Ashish Vaswani, Noam Shazeer, Niki Parmar, Jakob Uszkoreit, Llion Jones,
  Aidan~N Gomez, {\L}ukasz Kaiser, and Illia Polosukhin.
\newblock Attention is all you need.
\newblock {\em Advances in neural information processing systems}, 30, 2017.

\bibitem{wang2019combining}
Zhiyong Wang, Jinxiang Chai, and Shihong Xia.
\newblock Combining recurrent neural networks and adversarial training for
  human motion synthesis and control.
\newblock {\em IEEE transactions on visualization and computer graphics},
  27(1):14--28, 2019.

\bibitem{winkler2022questsim}
Alexander Winkler, Jungdam Won, and Yuting Ye.
\newblock Questsim: Human motion tracking from sparse sensors with simulated
  avatars.
\newblock {\em ACM TOG}, 2022.

\bibitem{yang2021lobstr}
Dongseok Yang, Doyeon Kim, and Sung-Hee Lee.
\newblock Lobstr: Real-time lower-body pose prediction from sparse upper-body
  tracking signals.
\newblock In {\em Comput. Graph. Forum}, volume~40, pages 265--275. Wiley
  Online Library, 2021.

\bibitem{ye2022neural3points}
Yongjing Ye, Libin Liu, Lei Hu, and Shihong Xia.
\newblock {Neural3Points: Learning to Generate Physically Realistic Full-body
  Motion for Virtual Reality Users}.
\newblock 2022.

\bibitem{yi2022physical}
Xinyu Yi, Yuxiao Zhou, Marc Habermann, Soshi Shimada, Vladislav Golyanik,
  Christian Theobalt, and Feng Xu.
\newblock Physical inertial poser (pip): Physics-aware real-time human motion
  tracking from sparse inertial sensors.
\newblock In {\em CVPR}, pages 13167--13178, 2022.

\bibitem{zhang2022motiondiffuse}
Mingyuan Zhang, Zhongang Cai, Liang Pan, Fangzhou Hong, Xinying Guo, Lei Yang,
  and Ziwei Liu.
\newblock Motiondiffuse: Text-driven human motion generation with diffusion
  model.
\newblock {\em arXiv preprint arXiv:2208.15001}, 2022.

\bibitem{zhou2019continuity}
Yi Zhou, Connelly Barnes, Jingwan Lu, Jimei Yang, and Hao Li.
\newblock On the continuity of rotation representations in neural networks.
\newblock In {\em CVPR}, pages 5745--5753, 2019.

\end{thebibliography}
}

\clearpage
\onecolumn
\appendix

\renewcommand*\thetable{\Roman{table}}
\renewcommand*\thefigure{\Roman{figure}}
\setcounter{table}{0}
\setcounter{figure}{0}

\section{Extra Ablation Experiments}

We first show extra ablation experiments of our method on AMASS~\cite{AMASS:ICCV:2019} dataset following the protocol proposed in~\cite{jiang2022avatarposer}. Then we show extra qualitative results and comparison between our method and the state-of-the-art method~\cite{jiang2022avatarposer}.

\paragraph{Additional Losses}
In addition to $\mathcal{L}_{dm}$, we explore three other geometric losses during the training like previous works~\cite{petrovich2021action, shi2020motionet}:

\begin{equation}
    \mathcal{L}_{pos} = \frac{1}{N} \sum_{i=1}^{N} \parallel \text{FK}(y^{i}_0) - \text{FK}(\hat{x}^{i}_0)\parallel_2^2
\end{equation}

\begin{equation}
\begin{array}{l}
    \mathcal{L}_{vel} = \frac{1}{N-1} \sum_{i=1}^{N-1} \parallel (\text{FK}(y^{i+1}_0) - \text{FK}(y^{i}_0)) - \\
    \quad \quad \quad \quad \quad \quad \quad \quad \quad (\text{FK}(\hat{x}^{i+1}_0) - \text{FK}(\hat{x}^{i}_0)) \parallel_2^2
\end{array}
\end{equation}

\begin{equation}
    \mathcal{L}_{foot} = \frac{1}{N-1} \sum_{i=1}^{N-1} \parallel (\text{FK}(y^{i}_0) - \text{FK}(\hat{x}^{i}_0)) \cdot m_i \parallel_2^2,
\end{equation}
where $\text{FK}(\cdot)$ is the forward kinematics function which takes local human joint rotations as input and outputs these joint positions in the global coordinate space. Here $y^{1:N}_0= x^{1:N}_0$ is the original data without noise. $\mathcal{L}_{pos}$ represents the position loss the of joints, $\mathcal{L}_{vel}$ represents the velocity loss of the joints in 3D space, and $\mathcal{L}_{foot}$ is the foot contact loss, which enforces static feet when there is no feet movement. $m_i \in \{0,1\}$ denotes the binary mask and equals to $0$ when the feet joints have zero velocity.

We train our model with different combinations of extra losses, setting their weights equal to $1$. As shown in Table~\ref{tab:ablation_losses}. In contrast to previous works~\cite{jiang2022avatarposer}, the extra geometric losses do not bring additional performance to our diffusion model. Our model can achieve good results when trained solely with the denoising objective function $\mathcal{L}_{dm}$ from Eq.~(4).%
We hypothesize that the lack of improvement in AGRoL's performance with additional losses is due to the intricacies of the reverse diffusion process. This process may not synergize effectively with extra geometrical losses without appropriate adjustments.

\begin{table}[h!]
    \centering
    \footnotesize
    \resizebox{0.88\textwidth}{!}{
    \begin{tabular}{ccc|cccccccc}
        \toprule
         $\mathcal{L}_{pos}$ & $\mathcal{L}_{vel}$ & $\mathcal{L}_{foot}$ & MPJRE & MPJPE & MPJVE & Hand PE & Upper PE & Lower PE & Root PE & Jitter\\
        \midrule
        \rowcolor{gray!9} & & & \bf 2.66 & \bf 3.71 & \bf 18.59 & \bf 1.31 & \bf 1.55 & \bf 6.84 & \bf 3.36 & \bf 7.26 \\
         & & \checkmark & 2.83 & 4.07 & 20.66 & 1.58 & 1.70 & 7.49 & 3.66 & 9.20 \\
        \checkmark &  &  & 2.81 & 4.06 & 21.85 & 1.75 & 1.73 & 7.43 & 3.72 & 12.16 \\ 
         \checkmark & \checkmark & & 2.73 & 3.92 & 20.55 & 1.72 & 1.68 & 7.15 & 3.52 & 10.16 \\
         \checkmark & \checkmark & \checkmark & 2.89 & 4.16 & 20.58 & 1.73 & 1.76 & 7.63 & 3.81 & 8.98 \\
        \bottomrule
    \end{tabular}}
    \vspace{-2mm}
    \caption{Ablation of the additional losses used during training.}
    \label{tab:ablation_losses}
\end{table}

\paragraph{Sampling Steps}
In Table~\ref{tab:ablation_sampling_steps} we ablate the number of sampling steps $T$ during training. Surprisingly, even when training with merely $10$ sampling steps, the model can achieve decent performance. Although we notice that the model converges to a worse local minimum when only a few sampling steps is used. To achieve the best results, more sampling steps is required. 

\begin{table}[!h]
    \centering
    \resizebox{0.72\textwidth}{!}{
    \begin{tabular}{l|ccccccc}
        \toprule
         \# Sampling Steps & MPJRE & MPJPE & MPJVE & Hand PE & Upper PE & Lower PE & Jitter\\
        \midrule
         10 & 2.69 & 3.70 & 19.41 & 1.47 & 1.55 & 6.80 & 7.63 \\
         100 & 2.65 & 3.62 & 18.74 & 1.33 & 1.52 & 6.66 & 6.71 \\
         \rowcolor{gray!9} 1000 (Ours) & 2.66 & 3.71 & 18.59 & 1.31 & 1.55 & 6.84 & 7.26 \\
        \bottomrule
    \end{tabular}
    } %
    \caption{Ablation of the number of sampling steps during training the AGRoL model. 
    The results become worse when the number of sampling steps is too small. More sampling steps is beneficial during training the network.}
    \label{tab:ablation_sampling_steps}
\end{table}

\paragraph{Input/Output length}
The proposed AGRoL model takes a sequence of sparse tracking signals as input and predicts the full body motion of the same length. In Table~\ref{tab:ablation_input_length} we ablate the input \& output length $N$ of the AGRoL model. Our model benefits from longer input sequences, especially decreasing the mean per joint velocity error and jitter. But the performance saturates after the length of $N=196$. In Table~\ref{tab:comparison_input_length} we further compare our method with AvatarPoser~\cite{jiang2022avatarposer} by varying its input length. Note that with longer input sequences our model can achieve significantly lower errors on velocity-related metrics like {MPJVE} and {jitter}, while AvatarPoser still has large {MPJVE} and {jitter} even with longer input length, thus failing to fully leverage the temporal information to generate smooth motions.

\begin{table}[!h]
    \centering
    \resizebox{0.72\textwidth}{!}{
    \begin{tabular}{l|ccccccc}
        \toprule
         Input \& Output Length & MPJRE & MPJPE & MPJVE & Hand PE & Upper PE & Lower PE & Jitter\\
        \midrule
         41 & 2.59 & 3.64 & 23.24 & 1.28 & 1.50 & 6.73 & 13.67 \\
         98 & 2.61 & 3.70 & 20.71 & 1.58 & 1.57 & 6.76 & 10.59 \\
         \rowcolor{gray!9} 196 (Ours) & 2.66 & 3.71 & 18.59 & 1.31 & 1.55 & 6.84 & 7.26 \\
         256 & 2.81 & 3.81 & 19.05 & 1.27 & 1.57 & 7.03 & 7.76 \\
        \bottomrule
    \end{tabular}
    } %
    \caption{Ablation of the input \& output length of the AGRoL model. Our model can benefit from larger input length.}
    \label{tab:ablation_input_length}
\end{table}

\begin{table}[!h]
    \centering
    \resizebox{0.78\textwidth}{!}{
    \begin{tabular}{l|c|ccccccc}
        \toprule
         Methods & Input Length & MPJRE & MPJPE & MPJVE & Hand PE & Upper PE & Lower PE & Jitter\\
        \midrule
        AvatarPoser~\cite{jiang2022avatarposer} & 41 & 3.08 & 4.18 & 27.70 & 2.12 & 1.81 & 7.59 & 14.49 \\
        AGRoL  & 41 & 2.59 & 3.64 & 23.24 & 1.28 & 1.50 & 6.73 & 13.67 \\
        \midrule
        AvatarPoser~\cite{jiang2022avatarposer} & 196 & 3.05 & 4.20 & 28.71 & 1.61 & 1.70 & 7.82 & 16.96 \\
        \rowcolor{gray!9} AGRoL (Ours) & 196 & 2.66 & 3.71 & 18.59 & 1.31 & 1.55 & 6.84 & 7.26 \\
        \bottomrule
    \end{tabular}
    } %
    \caption{Comparison between AGRoL and AvatarPoser~\cite{jiang2022avatarposer} while varying the number of input frames. Our method can benefit from longer inputs and generate smoother motion. In contrast, AvatarPoser fails to gain consistent improvement from longer input sequences and even degrades in some metrics, including MPJVE, Lower PE, and Jitter.}
    \label{tab:comparison_input_length}
\end{table}

\paragraph{Number of blocks in the MLP network}
In Table~\ref{tab:ablation_mlp_block} we evaluate the impact of varying the number of blocks (described in Sect.~3.2) %
in the MLP network. The model's performance consistently improves as more blocks are added. However, the performance gains approach a plateau when more than 12 blocks are used.

\begin{table}[h!]
    \centering
    \resizebox{0.85\textwidth}{!}{
    \begin{tabular}{l|ccccccccc}
        \toprule
         \#Blocks & \#Params & MPJRE & MPJPE & MPJVE & Hand PE & Upper PE & Lower PE & Root PE & Jitter\\
        \midrule
         2 & 2.08M & 4.54 & 7.39 & 34.38 & 3.10 & 3.13 & 13.55 & 7.28 & 22.03 \\
         6 & 4.35M & 2.89 & 4.12 & 19.51 & 1.38 & 1.69 & 7.62 & 3.72 & 6.29 \\
         \rowcolor{gray!9} 12 (Ours) & 7.48M & \textbf{2.66} & 3.71 & 18.59 & 1.31 & 1.55 & 6.84 & 3.36 & 7.26 \\
         24 & 14.53M & 2.73 & \textbf{3.61} & \textbf{18.33} & \textbf{1.08} & \textbf{1.50} & \textbf{6.65} & \textbf{3.28} & \textbf{7.23} \\
        \bottomrule
    \end{tabular}}
    \vspace{-2mm}
    \caption{Ablation study of the number of blocks in the proposed MLP network.}
    \label{tab:ablation_mlp_block}
\end{table}

\paragraph{Predicting noise}
Our diffusion model AGRoL follows \cite{ramesh2022hierarchical} and directly predicts the clean signal $\hat{x}_0^{1:N}$ in contrast to
the original Denoising Diffusion Probabilistic Model (DDPM) formulation~\cite{ho2020denoising}, where the model predicts residual noise $\epsilon_{\theta}(x_t, t)$ at every step. In this subsection, we further discuss the experiment presented in Table 3 of the main paper, where we implemented a version of AGRoL model (``AGRoL - {pred noise}") that predicts the residual noise $\epsilon_{\theta}(x_t, t)$. Similar to \cite{ramesh2022hierarchical}, we also find it better to predict the unnoised $\hat{x}_0^{1:N}$ directly, which is demonstrated by the results in Table~\ref{tab:ablation_arch_supp}. Since our simple MLP network (see Sect.~3.2 of the main paper) can already produce reasonable estimations of the full body motion using only one forward pass (see Table 1 in the main paper), we hypothesize that the DDPM formulation of Ramesh et al.~\cite{ramesh2022hierarchical} 
allows to exploit the full capacity of the network \emph{at every sampling step}, in contrast to the original formulation of~\cite{ho2020denoising}.

 \begin{table}[h!]
    \centering
    \footnotesize
    \begin{tabular}{l|ccccccccc}
        \toprule
         Method & MPJRE & MPJPE & MPJVE & Hand PE & Upper PE & Lower PE & Root PE & Jitter\\
        \midrule
         AGRoL - pred noise $\epsilon_{\theta}$ & 5.41 & 8.88 & 28.67 & 4.38 & 3.91 & 16.06 & 8.76 & 9.80 \\
         \rowcolor{gray!9} AGRoL (Ours) & \textbf{2.66} & \textbf{3.71} & \textbf{18.59} & \textbf{1.31} & \textbf{1.55} & \textbf{6.84} & \textbf{3.36} & \textbf{7.26} \\
        \bottomrule
    \end{tabular}
    \vspace{-2mm}
    \caption{Ablating different formulations of the diffusion model: Predicting clean signal directly (Ours) \emph{vs} predicting noise $\epsilon_{\theta}(x_t, t)$. The AGRoL model that learns to predict clean body motion at every diffusion step is substantially better in every metric.}
    \label{tab:ablation_arch_supp}
\end{table}

\section{Extra Datasets}
In addition to the AMASS~\cite{AMASS:ICCV:2019} dataset, we also evaluate the performance of our approach on AIST++~\cite{li2021ai} dataset. AIST++ dataset contains in total 5.1 hours of dancing movements performed by professional dancers. The dataset has 10 genres of dances, including some dances containing complicated movements like breakdancing, jazz etc. We follow the train/test splits proposed in~\cite{li2021ai}. The global rotation and translation of the hands and head are calculated using the SMPL human model~\cite{loper2015smpl} with the provided model parameters. Compared to the AMASS dataset, which contains mostly everyday life motions, the motions in the AIST++ dataset are much more diverse and challenging. As shown in Table~\ref{tab:aist_bench}, the AGRoL achieves superior performance in all the metrics and produces smoother motions compared to the AvatarPoser and the predictive MLP model. While there is still room for improvement on such a challenging dataset, the proposed AGRoL method significantly reduces the MPJVE, Jitter and lower body positional error (Lower PE) compared to the AvatarPoser.

\begin{table}[h!]
    \centering
    \footnotesize
    \begin{tabular}{l|ccccccccc}
        \toprule
         Method & MPJRE & MPJPE & MPJVE & Hand PE & Upper PE & Lower PE & Root PE & Jitter\\
        \midrule
         AvatarPoser & 4.37 & 9.11 & 97.24 & 4.31 & 3.32 & 17.47 & 8.11 & 65.18 \\
         MLP (Ours) & 3.63 & 7.33 & 74.90 & 3.86 & 2.69 & 14.03 & 5.48 & 47.16 \\
         \rowcolor{gray!9} AGRoL (Ours) & \bf 3.56 & \bf 6.83 & \bf 65.58 & \bf 2.17 & \bf 2.04 & \bf 13.74 & \bf 4.91 & \bf 41.95 \\
         \midrule
         GT & 0 & 0 & 0 & 0 & 0 & 0 & 0 & 30.48 \\
        \bottomrule
    \end{tabular}
    \caption{Comparison of our approach with the competitors on AIST++~\cite{li2021ai} dataset.}
    \label{tab:aist_bench}
\end{table}

\section{Extra Qualitative Results}
\begin{figure*}[t!]
    \centering
    \begin{subfigure}[b]{0.33\linewidth}
    \caption{GT}
    \end{subfigure}
    \begin{subfigure}[b]{0.33\linewidth}
    \caption{AvatarPoser~\cite{jiang2022avatarposer}}
    \end{subfigure}
    \begin{subfigure}[b]{0.33\linewidth}
    \caption{AGRoL (Ours)}
    \end{subfigure}
    
    \begin{subfigure}[b]{0.97\linewidth}
    \includegraphics[width=1.0\linewidth]{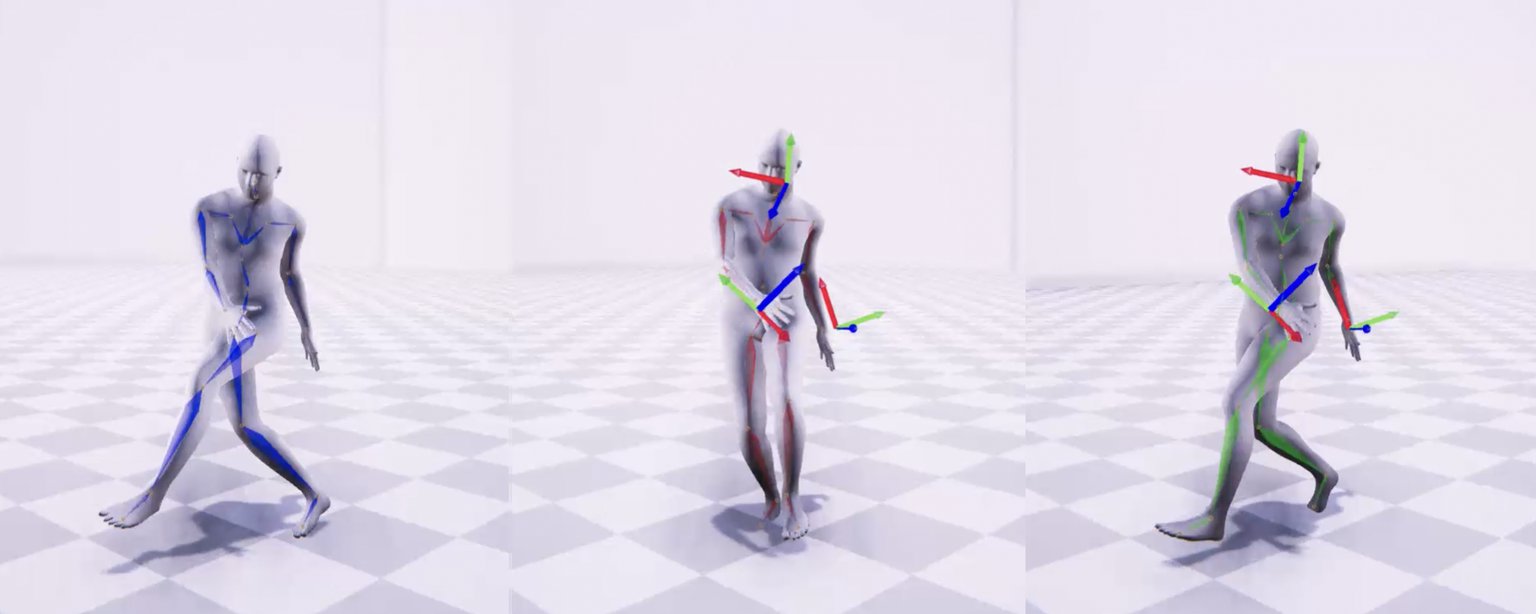} %
    \end{subfigure}
    \begin{subfigure}[b]{0.97\linewidth}
    \includegraphics[width=1.0\linewidth,height=0.4\linewidth]{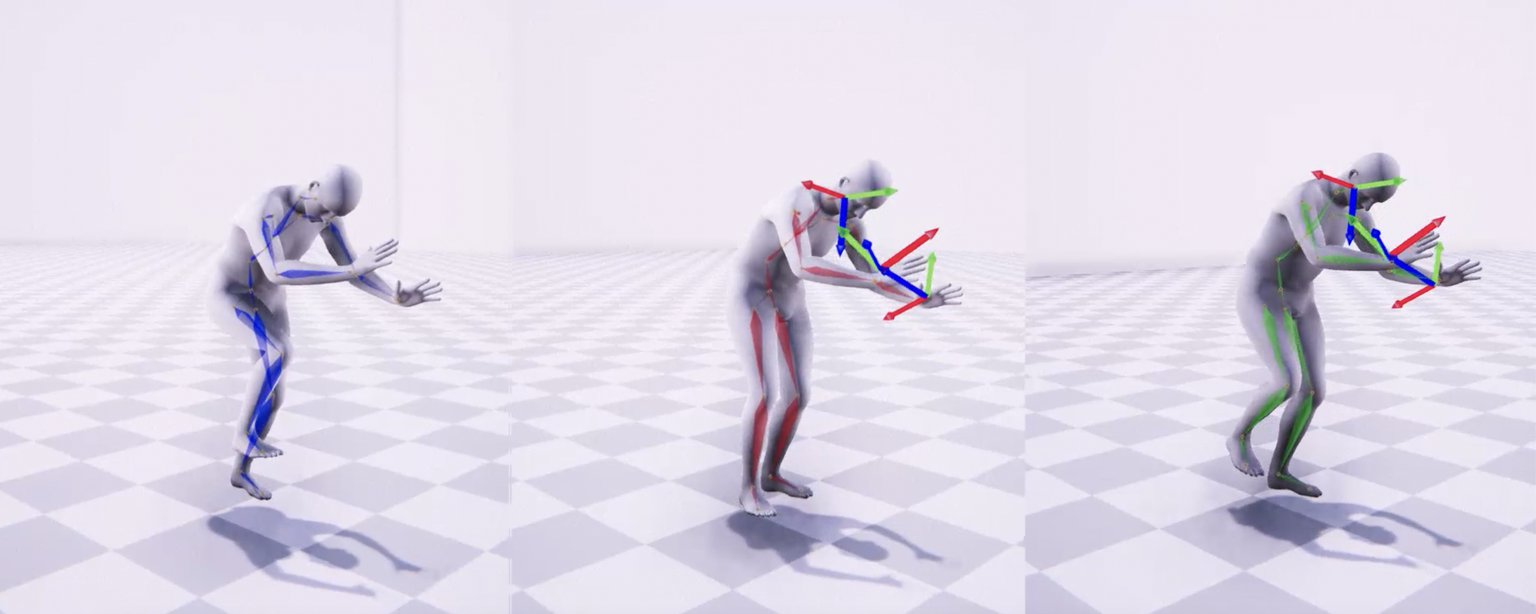}
    \end{subfigure}
    \begin{subfigure}[b]{0.97\linewidth}
    \includegraphics[width=1.005\linewidth,height=0.4\linewidth]{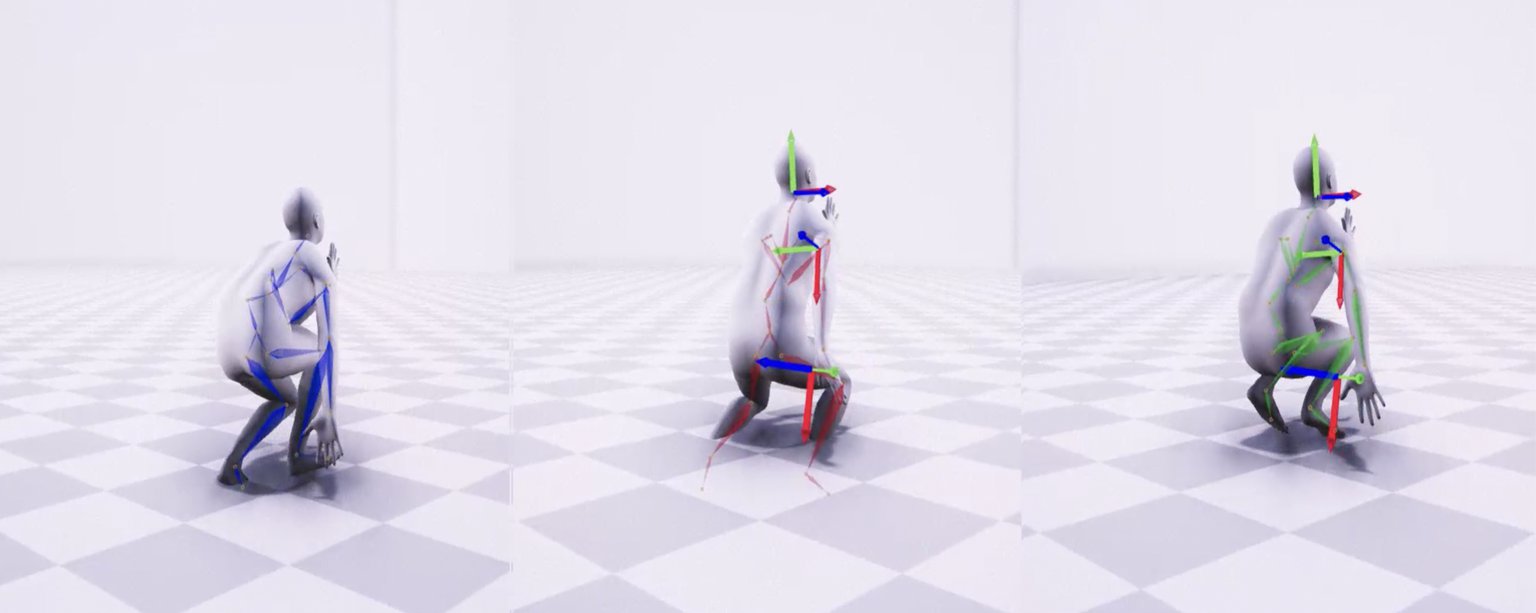}
    \end{subfigure}
    \vspace{-3mm}
    \caption{\textbf{Qualitative comparison} on test sequences from AMASS dataset. We visualize the predicted skeletons and human body meshes: \textbf{(a)} Ground truth skeletons in \textcolor{traj_blue}{blue},  \textbf{(b)} AvatarPoser predictions in \textcolor{traj_red}{red}, \textbf{(c)}  AGRoL predictions in \textcolor{traj_green}{green}. It can be seen that our predicted motion is more accurate compared to the predictions of AvatarPoser, especially in the lower body. RGB axes illustrate the location and orientation of the head and hands provided as input to the models. \emph{Please refer to our \href{https://dulucas.github.io/agrol/}{project page} for more qualitative results.}}
    \label{fig:qualitative_results}
\end{figure*}

In Figure~\ref{fig:qualitative_results} we show extra qualitative comparisons between our method and AvatarPoser~\cite{jiang2022avatarposer}. Please refer to the videos in the supplementary material\footnote{\label{note_web_page}\url{https://dulucas.github.io/agrol/}.} for more qualitative results. As shown in the video, our method reconstructs the full body poses more accurately, it can generate smoother motions and alleviate the jittering issue compared to AvatarPoser.

\paragraph{Real inputs}
Additionally, in the supplementary material\footref{note_web_page}, we include a video showcasing AGRoL inference on real VR inputs from the Quest HMD. Our model exhibits a reasonable generalization to the real-world inputs and is capable of generating smooth and precise motions.

\begin{figure*}[t!]
    \centering
    \begin{subfigure}[b]{0.33\linewidth}
    \caption{GT}
    \end{subfigure}
    \begin{subfigure}[b]{0.33\linewidth}
    \caption{AvatarPoser~\cite{jiang2022avatarposer}}
    \end{subfigure}
    \begin{subfigure}[b]{0.33\linewidth}
    \caption{AGRoL (Ours)}
    \end{subfigure}
    
    \begin{subfigure}[b]{0.94\linewidth}
    \includegraphics[width=1.0\linewidth,height=0.4\linewidth]{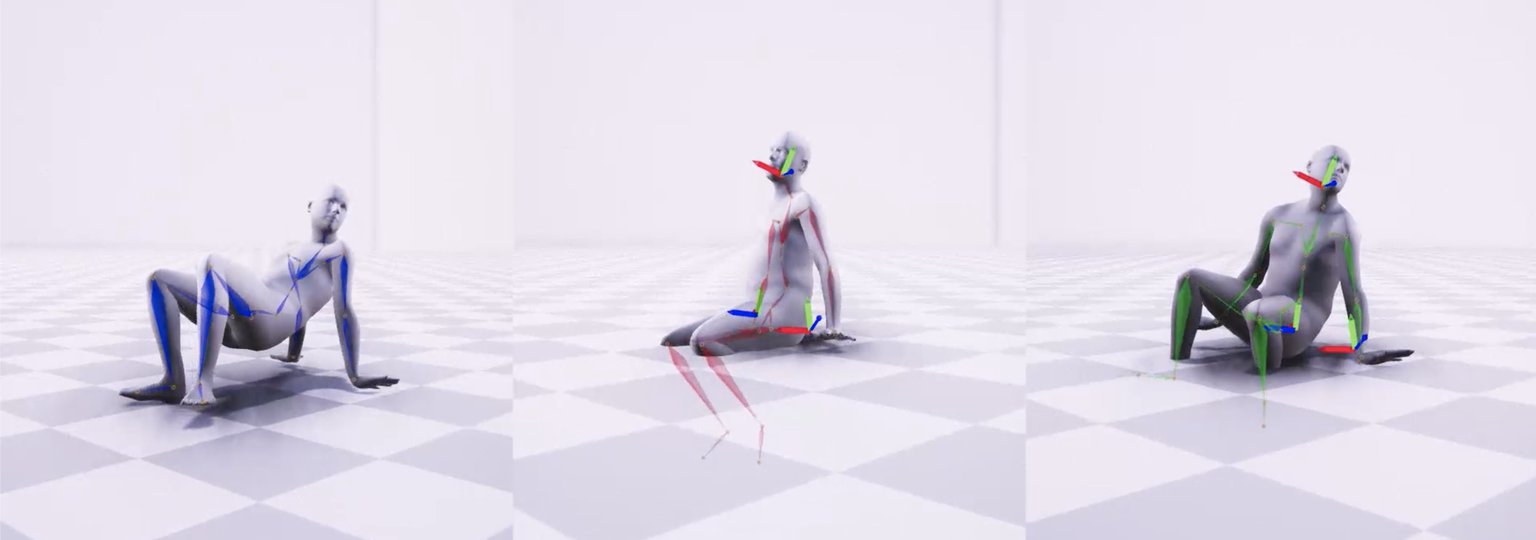}
    \end{subfigure}
    \begin{subfigure}[b]{0.94\linewidth}
    \includegraphics[width=1.0\linewidth,height=0.4\linewidth]{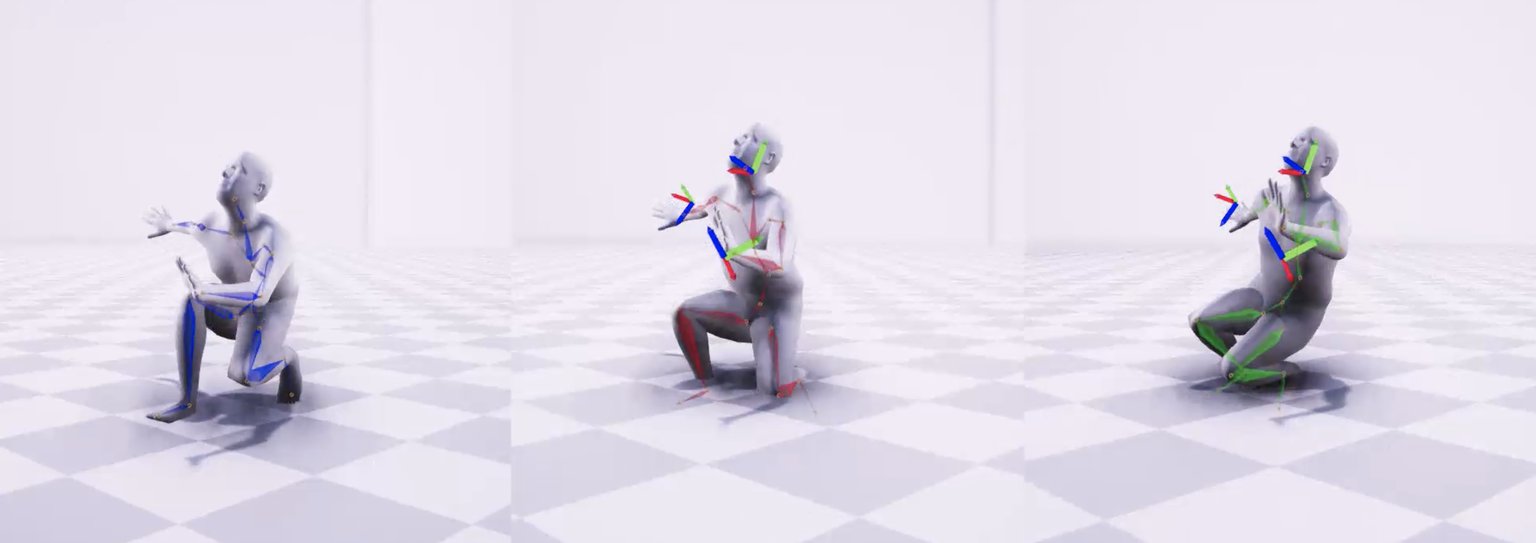}
    \end{subfigure}
    \begin{subfigure}[b]{0.94\linewidth}
    \includegraphics[width=1.0\linewidth,height=0.4\linewidth]{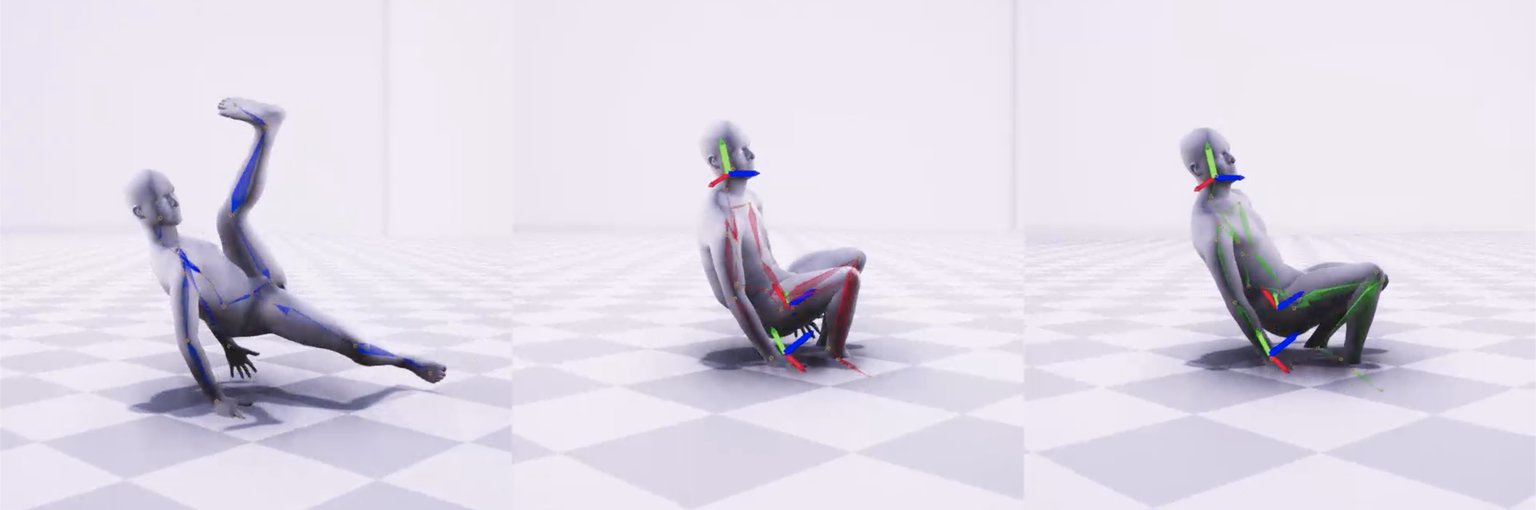}
    \end{subfigure}
    \caption{\textbf{Failure test cases.} We visualize the predicted skeletons and human body meshes: \textbf{(a)} Ground truth skeletons in \textcolor{traj_blue}{blue},  \textbf{(b)} AvatarPoser predictions in \textcolor{traj_red}{red}, \textbf{(c)}  AGRoL predictions in \textcolor{traj_green}{green}. The models were trained on the subset of AMASS~\cite{AMASS:ICCV:2019} following the protocol of~\cite{jiang2022avatarposer}. In the figures, the RGB axes indicate the position and orientation of the head and hands, which are used as input to the models. We can observe that our method struggles with (i) irregular poses that were not well-represented in the training set, resulting in inaccurate poses and floor penetration, and (ii) when the lower body pose is not strongly correlated with the upper body, as seen in the break dance motion in the bottom row. To address these failure cases, we could consider increasing the diversity of training motions and incorporating additional physical constraints.}
    \label{fig:failures}
\end{figure*}

\paragraph{Failure cases}
We also demonstrate failure cases for the proposed AGRoL model in Figure~\ref{fig:failures}.
We can see that our method fails when we test it on irregular poses, that were not well covered in the training set, or when the lower body pose does not have strong correlation with the upper body. For example, during the break dance motion (Fig.~\ref{fig:failures}, bottom row) the upper body may stay static, while legs move which makes it very challenging to predict legs accurately.
Increasing the size and diversity of the training set plus incorporating extra physical or geometrical priors to prevent floor penetration could be a potential solution for the failure cases. We plan to further investigate it in future work.

\end{document}